# SPINEX_ Symbolic Regression: Similarity-based Symbolic Regression with Explainable Neighbors Exploration

M.Z. Naser[1,2], Ahmed Z. Naser[3]

[1]School of Civil & Environmental Engineering and Earth Sciences (SCEEES), Clemson University, USA
[2]Artificial Intelligence Research Institute for Science and Engineering (AIRISE), Clemson University, USA
E-mail: mznaser@clemson.edu, Website: www.mznaser.com
[3]Department of Mechanical Engineering, University of Manitoba, Canada, E-mail: a.naser@umanitoba.ca

**Abstract**

This article introduces a new symbolic regression algorithm based on the SPINEX (Similarity-based Predictions with Explainable Neighbors Exploration) family. This new algorithm (SPINEX_SymbolicRegression) adopts a similarity-based approach to identifying high-merit expressions that satisfy accuracy- and structural similarity metrics. We conducted extensive benchmarking tests comparing SPINEX_SymbolicRegression to over 180 mathematical benchmarking functions from international problem sets that span randomly generated expressions and those based on real physical phenomena. Then, we evaluated the performance of the proposed algorithm in terms of accuracy, expression similarity in terms of presence operators and variables (as compared to the actual expressions), population size, and number of generations at convergence. The results indicate that SPINEX_SymbolicRegression consistently performs well and can, in some instances, outperform leading algorithms. In addition, the algorithm's explainability capabilities are highlighted through in-depth experiments.

## 1.0 Introduction

With the continued rise of machine learning (ML), symbolic regression has emerged as a novel approach for uncovering the underlying mathematical relationships [1]. The interest in symbolic regression stems from the fact that, unlike traditional regression techniques that rely on predefined model structures, symbolic regression autonomously discovers the form and parameters of the mathematical expressions that best describe the data. This apparent flexibility enables the identification of unique expressions (i.e., models) that may not be apparent through conventional modeling approaches [2].

Symbolic regression often leverages evolutionary algorithms (e.g., genetic programming) to explore the search space of potential mathematical expressions and operators [3]. Such algorithms iteratively evolve candidate solutions through processes analogous to natural selection and attempt to optimize both the structure and coefficients of mathematical models [4]. This evolutionary paradigm allows for the simultaneous optimization of multiple aspects of the model, including non-linear relationships and interactions between variables that can be challenging to capture with traditional regression methods. Furthermore, the recent advancements in symbolic regression that can accommodate multi-expression models and domain-specific knowledge through priors/constraints have significantly enhanced the performance and applicability of this technique [5,6].

The significance of symbolic regression lies in its ability to produce explicit mathematical descriptions of the data and phenomena at hand. On the other hand, traditional ML models, such as XGBoost or deep neural networks, offer high predictive accuracy at the expense of interpretability, while symbolic regression yields accurate and transparent models [7]. Symbolic regression addresses the rising demands for interpretability by providing inherently understandable and justifiable models to foster trust and facilitate their adoption. Furthermore, the explicit nature of the symbolic models enables easier validation and verification, which relies on examining how such models adhere to known scientific principles and constraints [8].

The above dual advantage is particularly valuable in science and engineering. This is because these disciplines emphasize understanding the underlying mechanisms as crucial as making precise predictions. For instance, in fields like physics, symbolic regression can reveal fundamental laws and interactions that can lay the ground for theoretical advancements [9]. More specifically, symbolic regression has been employed to rediscover classical laws from empirical data, validate theoretical models, and uncover new physical relationships [10]. In addition, engineering disciplines utilize symbolic regression for system identification, control strategy development, and optimization of design parameters [11]. In more recent attempts, symbolic regression has been reported to provide a means for the automated generation of hypotheses and the derivation of mathematical models from experimental data [12].

The focus of this study is to capitalize on the concept of similarity as the driving force behind symbolic regression. This focus has not been fully explored or investigated [13,14]. Yet, the concept of similarity can play a pivotal role in enhancing the robustness of symbolic regression. More specifically, by leveraging similarity measures, symbolic regression algorithms can more effectively navigate the search space, identify relevant patterns, and ensure the generation of meaningful and generalizable models. Further, similarity in symbolic regression can utilize similarity metrics to evaluate the resemblance between candidate models and the underlying data structure via metrics such as Euclidean distance, cosine similarity, etc. Thus, incorporating similarity can allow symbolic regression algorithms to prioritize models that fit the data well and exhibit structural similarities to known or hypothesized relationships.

For instance, data clustering can identify subsets of the dataset that share common characteristics, allowing the algorithm to tailor the search for mathematical expressions that capture these localized patterns. Similarly, incorporating similarity-based constraints can guide the evolutionary process toward models that adhere to predefined structural or functional similarities. These constraints may derive from domain-specific knowledge, such as known physical laws or symmetry properties, to ensure that the generated models are accurate and consistent with established theoretical principles. Similarity-based pattern recognition techniques can also be employed to identify recurring mathematical structures (and/or operators) within the data to construct models that inherently align with the intrinsic data dynamics [15]. For example, crossover and mutation operators can be tailored to maintain structural or functional similarities between parent and offspring models, ensuring that beneficial traits are retained and propagated across generations [16].

The above presents a case for this work wherein a new symbolic regression algorithm is proposed. This algorithm builds upon the success of the recently developed algorithmic architecture, SPINEX (Similarity-based Predictions with Explainable Neighbors Exploration), which adopts the concept of similarity and explainability. More specifically, we present the SPINEX SymbolicRegressor and its mechanisms; then, we examine the algorithm's performance on 182 diverse benchmarking functions.

## 2.0 Approaches to Symbolic Regression

There are a number of methodologies for symbolic regression. This section briefly reviews the most prominent approaches, including genetic algorithms (GAs), genetic programming (GP), deep learning-based methods, hybrid techniques, Bayesian approaches, and multi-objective frameworks.

### 2.1 Genetic algorithms (GAs)

The philosophy of natural selection and genetics inspires genetic algorithms. In this context, GAs encode potential mathematical models as chromosomes that can be represented as fixed-length strings or binary sequences. These chromosomes undergo genetic operations such as crossover (recombination) and mutation to explore the search space of possible expressions. Then, GAs incorporate fitness functions to evaluate candidate expressions based on how well they fit the data, as well as to guide the evolutionary process toward more accurate expressions. GAs often require careful tuning and may struggle with representing complex, variable-length expressions. Some of the notable researchers who pioneered GAs include Koza [17] and Holland [18].

### 2.2 Genetic programming (GP)

Genetic programming extends the principles of GAs to the domain of program synthesis by allowing for the evolution of tree-structured representations of mathematical expressions. In GP, each individual in the population is typically represented as a parse tree, where internal nodes correspond to mathematical operators (e.g., addition, multiplication) and leaf nodes represent variables or constants. This tree-based structure provides the flexibility to represent and manipulate complex, hierarchical expressions naturally. GP employs genetic operations similar to GAs but is better suited to handling the variable-length and nested nature of mathematical expressions. Advanced GP techniques incorporate mechanisms such as subtree crossover, mutation, and hierarchical fitness evaluation to enhance evolved models' diversity and quality, as seen in [19,20].

### 2.3 Deep learning-based methods

In this form of symbolic regression, neural network architectures, particularly those designed for sequence generation and transformation, are employed to predict mathematical expressions that fit the data. For example, recurrent neural networks (RNNs) and transformer models can be trained to generate symbolic expressions by treating the problem as a sequence-to-sequence task. These models leverage their ability to capture complex patterns and dependencies within data to arrive at diverse operators and expressions. On a similar front, reinforcement learning approaches have been integrated with deep learning to guide the generation of expressions based on reward signals related to model fitness, thereby optimizing the search process in a data-driven manner [21,22].

### 2.4 Bayesian symbolic regression

Bayesian approaches introduce probabilistic frameworks to symbolic regression and, therefore, can enable the incorporation of prior knowledge and the quantification of uncertainty in the discovered expressions [23]. Such an approach typically employs probabilistic models to represent the distribution over possible mathematical expressions. In addition, such an approach is capable of updating beliefs about the model structure and parameters based on observed data. Techniques such as Markov Chain Monte Carlo (MCMC) and variational inference are utilized to sample from or approximate the posterior distribution to facilitate the exploration of the search space in a principled manner. This probabilistic approach enhances the robustness and interpretability of symbolic regression by providing confidence measures and allowing for the integration of domain-specific priors.

*2.5 Hybrid and ensemble methods*

More recently, hybrid and ensemble methods have been developed for symbolic regression to capitalize on the strengths of multiple approaches [24]. These methods combine genetic programming with local optimization algorithms, such as gradient descent, to fine-tune the parameters of evolved models [25]. In addition, the incorporation of local algorithms can aid in improving the accuracy and convergence speed of the symbolic regressors. In some architectures, ensemble methods can aggregate multiple symbolic models to enhance predictive performance and generalization capabilities.

*2.6 Multi-objective optimization*

Multi-objective symbolic regression methods can simultaneously optimize multiple criteria, such as expression accuracy and complexity [26]. By treating these objectives as competing factors, multi-objective optimization techniques, including Pareto-based approaches, seek to identify a set of non-dominated solutions that offer trade-offs between the objectives. Such a method enables the discovery of models that balance interpretability and predictive power, catering to applications where both factors are critical.

### 3.0 Description of the SPINEX SymbolicRegressor

*3.1 General description*

The SPINEX_SymbolicRegression algorithm presents a sophisticated symbolic regression approach that leverages similarity and genetic programming as core techniques. This algorithm initializes a diverse population of symbolic expressions composed of variables, constants, and a variety of unary and binary operators, including trigonometric functions, exponential functions, and other mathematical operations. Some of the key features of SPINEX_SymbolicRegression include a time constraint limit to monitor computational. The inclusion of an actual_function allows the algorithm to compare evolved expressions against a known ground truth, enhancing validation capabilities. Further, users can inject custom expressions (user_expression) into the initial population, guiding the evolutionary process based on domain knowledge.

The algorithm maintains genetic diversity and prevents premature convergence via similarity_threshold, which quantifies the structural similarity between expressions using tree edit distance metrics. In addition, the algorithm also uses dynamic_similarity_threshold to adjust this threshold during evolution in order to respond to the diversity levels within the population. An adaptive elite selection mechanism (dynamic_elite) adjusts the proportion of top-performing expressions carried over to subsequent generations. This dynamic adjustment is based on the observed diversity and convergence rates, balancing exploration and exploitation within the search space.

The relevance_metric assesses variable importance using statistical measures like correlation and mutual information as means to inform variable selection and guiding the inclusion of relevant variables in expressions. More specifically, the algorithm supports different variable inclusion strategies (variable_inclusion_strategy), such as guided or probabilistic, which dictate how variables are incorporated into expressions. The force_all_variables option ensures that all input variables are represented in the final expression if required by the user. A number of evolutionary operators control the mutation method (with strategies like replace, add, remove) and crossover, which recombine subexpressions from parent expressions. The algorithm carefully manages expression complexity, applying a dynamic_complexity_penalty to balance fitness with model simplicity, thereby preventing overfitting.

In addition to the above, the proposed algorithm also incorporates several features that enhance the explainability of the symbolic expressions it produces. At a basic level, by setting force_all_variables to True, the algorithm ensures that all input variables are included in the final expression, which can be particularly useful when domain knowledge dictates that certain variables are essential to the model. Then, the variable_inclusion_strategy parameter allows for different methods of including variables:
- Guided strategy: assesses the relevance of each variable using metrics like correlation and/or mutual information. Variables deemed more relevant are prioritized during the expression generation process.
- Probabilistic strategy: Variables are included based on probabilistic rules, which can introduce expression diversity while promoting important variables' inclusion.

As seen above, the user_expression parameter allows users to inject domain-specific knowledge directly into the algorithm. The algorithm can incorporate known relationships or constraints into the evolution process by providing a user-defined mathematical expression. This guides the search towards more meaningful expressions and warrants the final expression to align with established theoretical frameworks. The explainability_level parameter allows users to control the number of explanations about the evolution process. Two levels are offered, basic or advanced, which can direct the algorithm to output additional information, such as:
- Variable Importance: Detailed reports on how variables influence the evolution of expressions.
- Evolutionary Process: Insights into how expressions mutate and crossover over generations.

*3.2 Detailed description*

A more detailed description of SPINEX_SymbolicRegression's methods and functions is provided herein.

**Classes and Exceptions**
**TimeoutMixin**
A mixin class to provide timeout functionality to classes that inherit from it.
- check_timeout(self)

- o **Description:** Checks if the elapsed time since start_time exceeds max_time. If so, raises a TimeoutException.
- o **Mathematical Notation:** if elapsed=current_time−start_time>max_time⇒raise TimeoutException

**TimeoutException**
Custom exception raised when a timeout condition is met.

## Standalone Functions

### timeout_handler(timeout_duration, function, *args, **kwargs)
Executes a function with a specified timeout. If the function does not complete within timeout_duration, a TimeoutException is raised.
- **Parameters:**
  - o timeout_duration (float): Maximum allowed time in seconds.
  - o function (callable): The target function to execute.
- **Behavior:**
  - o Executes function(*args, **kwargs) in a separate thread.
  - o If execution exceeds timeout_duration, raises TimeoutException.
- **Mathematical Representation:** if execution_time>timeout_duration⇒raise TimeoutException

### set_seed(seed_value=42)
Initializes the random number generators for NumPy and Python's random module to ensure reproducibility.

### cached_get_expression_depth_helper(expr)
Calculates the depth of a symbolic expression using recursion and caches results for efficiency.
- **Parameters:**
  - o expr (SymPy expression): The expression whose depth is to be calculated.
- **Returns:** Integer representing the maximum depth of expr.
- **Mathematical Definition:** depth(expr)=1+max(depth(arg) | arg ∈ expr.args)

### clear_all_lru_caches()
Clears all Least Recently Used (LRU) caches used in the algorithm to free memory and reset cached computations.

## SPINEX_SymbolicRegression Class

### Initialization (__init__)
The __init__ encapsulates the essential elements and sets up the operational parameters of SPINEX. The method signature is as follows:

```
class SPINEX_SymbolicRegression(TimeoutMixin):
   def __init__(self, data, target, actual_function, user_expression=None, max_depth=None, population_size=50,
        generations=50, similarity_threshold=0.7, dynamic_similarity_threshold=True, n_jobs=1,
        explainability_level='none', plot_results=True, early_stopping_metric=None, early_stopping_value=None,
        patience=0, force_all_variables=False, variable_inclusion_strategy='guided', logger=None,
        last_resort=False, elite_percentage=0.05, dynamic_elite=True, max_time=None, relevance_metric='both'):
     self.max_time = max_time
     self.start_time = time.time()
     self._expression_depth_cache = {}
     self._evaluate_expression_cache = {}
     self._subexpressions_cache = {}
     self._tree_edit_distance_cache = {}
     self.tree_edit_distance_cache_lock = threading.Lock()
     self.similarity_cache = {}
     self.similarity_cache_lock = threading.Lock()
     self.lambdified_cache = {}
     self.data = data
```

```python
        self.target = target
        self.actual_function = actual_function
        self.force_all_variables = force_all_variables
        self.variable_inclusion_strategy = variable_inclusion_strategy
        self.variables = {str(var): sp.Symbol(var) for var in list(data.columns)}
        if logger is None:
            self.logger = Logger()
        else:
            self.logger = logger
        base_depth = 2
        if max_depth is None:
            self.max_depth = base_depth + (len(self.variables) if self.force_all_variables else 2)
            self.logger.debug(f"max_depth set to {self.max_depth} based on {'all variables' if self.force_all_variables else 'base depth + 2'}")
        else:
            min_depth = base_depth + (len(self.variables) if self.force_all_variables else 2)
            if max_depth < min_depth:
                self.logger.warning(f"Provided max_depth={max_depth} is too low. Setting to minimum required: {min_depth}.")
                self.max_depth = min_depth
            else:
                self.max_depth = max_depth
                self.logger.debug(f"max_depth provided: {self.max_depth}")
        self.initial_max_depth = self.max_depth
        self.logger.debug(f"initial_max_depth set to {self.initial_max_depth}")
        self.relevance_metric = relevance_metric
        self.relevance_scores = self.assess_variable_relevance(relevance_metric=self.relevance_metric)
        self.sorted_vars = sorted(
            self.variables.values(),
            key=lambda var: self.relevance_scores.get(var, 0),
            reverse=True
        )
        self.logger.debug(f"Variables sorted by relevance: {self.sorted_vars}")
        self.user_expression = user_expression
        if self.user_expression:
            try:
                self.user_expression_sym = sp.sympify(self.user_expression)
                if self.get_expression_depth(self.user_expression_sym) > self.max_depth:
                    self.logger.warning("User expression exceeds max_depth. It will be truncated or modified.")
                    self.user_expression_sym = self.truncate_expression(self.user_expression_sym)
                allowed_symbols = set(self.variables.values())
                if not self.user_expression_sym.free_symbols.issubset(allowed_symbols):
                    self.logger.warning("User expression contains variables not present in the data. They will be removed.")
                    self.user_expression_sym = self.user_expression_sym.subs(
                        {var: 0 for var in self.user_expression_sym.free_symbols if var not in allowed_symbols}
                    )
            except Exception as e:
                self.logger.error(f"Invalid user_expression provided: {e}")
                self.user_expression_sym = None
```

```python
        else:
            self.user_expression_sym = None
    self.population_size = population_size
    self.generations = generations
    self.similarity_threshold = similarity_threshold
    self.initial_similarity_threshold = similarity_threshold
    self.dynamic_similarity_threshold = dynamic_similarity_threshold
    self.elite_percentage = elite_percentage
    self.dynamic_elite = dynamic_elite
    if not (0 < self.elite_percentage <= 1):
        raise ValueError("elite_percentage must be a float between 0 and 1")
    self.n_jobs = n_jobs
    self.explainability_level = explainability_level
    self.plot_results = plot_results
    self.early_stopping_metric = early_stopping_metric
    valid_metrics = ['fitness', 'r2', 'mse']
    if self.early_stopping_metric is not None:
        self.early_stopping_metric = self.early_stopping_metric.lower()
        if self.early_stopping_metric not in valid_metrics:
            raise ValueError(f"Invalid early_stopping_metric: {early_stopping_metric}. Choose from {valid_metrics}.")
        if early_stopping_value is None:
            raise ValueError("early_stopping_value must be provided when early_stopping_metric is set.")
    self.early_stopping_value = early_stopping_value
    self.patience = patience
    self.fitness_tolerance = 1e-6
    self.epsilon = 1e-10
    self.lambdified_cache = {}
    self.population = []
    self.best_expression = None
    self.baseline_mse = mean_squared_error(self.target, np.mean(self.target) * np.ones_like(self.target))
    self.expression_cache = {}
    self.max_cache_size = 1000
    self.diversity_weight = 0.3
    self.mutation_types = ['replace', 'add', 'remove']
    valid_explain_levels = ['none', 'basic', 'advanced']
    if self.explainability_level.lower() not in valid_explain_levels:
        raise ValueError(f"Invalid explainability_level: {self.explainability_level}. "
                         f"Choose from {valid_explain_levels}.")
    self.unary_operators = [
        sp.sin, sp.cos,
        self.safe_exp, self.safe_log,
        self.safe_sqrt, self.safe_atan,
        self.safe_acos, self.safe_asin,
        self.safe_sinh, self.safe_cosh,
        self.safe_tanh
    ]
    self.binary_operators = [
        sp.Add,
        lambda x, y: x - y,
```

```
        sp.Mul,
        self.safe_div,
        self.safe_max,
        self.safe_min,
        self.safe_pow
    ]
    self.constant_choices = [sp.Integer(i) for i in range(-10, 11)] + \
                    [sp.pi, sp.E, sp.Rational(1, 2), sp.Rational(1, 3),
                     sp.Rational(1, 4), sp.Rational(1, 5), sp.sqrt(2), sp.sqrt(3)]
    self.variable_values = list(self.variables.values())
    self.variable_and_constant_choices = self.variable_values + self.constant_choices
    self.all_variables_set = set(self.variables.values())
    self.lambdify_modules = ['numpy', {
        'Max': np.maximum,
        'Min': np.minimum,
        'sin': np.sin,
        'cos': np.cos,
        'tan': np.tan,
        'asin': np.arcsin,
        'acos': np.arccos,
        'atan': np.arctan,
        'sinh': np.sinh,
        'cosh': np.cosh,
        'tanh': np.tanh,
        'exp': np.exp,
        'sqrt': np.sqrt,
        'Abs': np.abs,
        'ceiling': np.ceil,
        'floor': np.floor
    }]
    self.unit_array = np.array([1.0])
    self.num_variables = len(self.variables)
    self.sample_value_arrays = [self.unit_array] * self.num_variables
    self.last_resort = last_resort
    self.last_resort_activated = False
    self.logger.info("SPINEX_SymbolicRegression initialized with parameters:")
    self.logger.info(f"max_depth={max_depth}, population_size={population_size}, generations={generations}, "
            f"similarity_threshold={similarity_threshold}, n_jobs={n_jobs}, "
            f"explainability_level={explainability_level}, plot_results={plot_results}, "
            f"early_stopping_fitness={early_stopping_value}, patience={patience}, "
            f"force_all_variables={force_all_variables}, variable_inclusion_strategy={variable_inclusion_strategy}")
```

More specifically:
1. **Data and Target**
    - data: A pandas DataFrame containing the input features.
    - target: The target variable that the regression aims to predict.
    - actual_function: The true underlying function (if known) for comparison or validation purposes.
2. **Expression and Population Control**

- o   user_expression (optional): A user-provided symbolic expression to guide or constrain the regression process.
- o   max_depth (optional): The maximum depth of the expression trees, controlling the complexity of the symbolic expressions.
- o   population_size (default=50): The number of candidate expressions in each generation.
- o   generations (default=50): The number of evolutionary cycles to perform.

3. **Similarity and Diversity Management**
   - o   similarity_threshold (default=0.7): Threshold to determine the similarity between expressions, aiding in maintaining diversity.
   - o   dynamic_similarity_threshold (default=True): Whether to adjust the similarity threshold dynamically during evolution.

4. **Parallelism and Performance**
   - o   n_jobs (default=1): Number of parallel jobs to run, facilitating multi-core processing.

5. **Explainability and Visualization**
   - o   explainability_level (default='none'): Level of explainability for the resulting model ('none', 'basic', 'advanced').
   - o   plot_results (default=True): Whether to generate plots of the results.

6. **Early Stopping Mechanism**
   - o   early_stopping_metric (optional): Metric to monitor for early stopping ('fitness', 'r2', 'mse').
   - o   early_stopping_value (optional): The target value for the early stopping metric.
   - o   patience (default=0): Number of generations to wait for improvement before stopping.

7. **Variable and Expression Handling**
   - o   force_all_variables (default=False): Whether to force the inclusion of all variables in the expressions.
   - o   variable_inclusion_strategy (default='guided'): Strategy for including variables ('guided' suggests a guided approach based on relevance).

8. **Logging and Monitoring**
   - o   logger (optional): Custom logger for tracking the regression process.

9. **Additional Controls**
   - o   last_resort (default=False): Whether to activate a last-resort mechanism if initial attempts fail.
   - o   elite_percentage (default=0.05): Percentage of top-performing expressions retained in each generation.
   - o   dynamic_elite (default=True): Whether to adjust the elite percentage dynamically.
   - o   max_time (optional): Maximum time allowed for the regression process.
   - o   relevance_metric (default='both'): Metric for assessing variable relevance ('both' likely combines multiple criteria).

## Initialization Steps

The constructor initializes various attributes and sets up the necessary environment for symbolic regression:

1. **Time Management**
   - o   self.max_time: Stores the maximum allowed time for the regression.
   - o   self.start_time: Captures the start time of the regression process using time.time().

2. **Caching Mechanisms**
   - o   Several dictionaries (_expression_depth_cache, _evaluate_expression_cache, _subexpressions_cache, _tree_edit_distance_cache, similarity_cache, lambdified_cache) are initialized to cache computations, enhancing performance by avoiding redundant calculations.
   - o   Threading locks (tree_edit_distance_cache_lock, similarity_cache_lock) are established to manage concurrent access to shared caches, ensuring thread safety.

3. **Data and Variables Setup**
   - o   self.data, self.target, self.actual_function: Store the input data, target variable, and the actual function, respectively.
   - o   self.variables: Creates a dictionary mapping each column in the data to a SymPy symbol, facilitating symbolic manipulation.
   - o   self.num_variables: Counts the number of input variables.

4. **Logging Setup**

- o If no custom logger is provided, a default Logger instance is created.
- o Logs are used extensively to debug and monitor the regression process, providing insights into parameter settings and internal states.

5. **Depth Configuration**
   - o Determines self.max_depth, the maximum depth of expression trees. If not provided, it defaults based on the number of variables and a base depth.
   - o Validates the provided max_depth, ensuring it meets a minimum requirement based on the number of variables if force_all_variables is set.

6. **Variable Relevance Assessment**
   - o self.relevance_metric: Specifies the metric used to assess the relevance of each variable.
   - o self.relevance_scores: Stores the relevance scores for each variable, determined by self.assess_variable_relevance.
   - o self.sorted_vars: Lists variables sorted by their relevance, guiding the inclusion strategy in expressions.

7. **User Expression Handling**
   - o If a user_expression is provided, it is parsed and converted into a SymPy expression (self.user_expression_sym).
   - o Validates the depth of the user expression, truncating or modifying it if it exceeds max_depth.
   - o Ensures that the user expression only contains variables present in the data, removing or substituting any extraneous variables.

8. **Genetic Algorithm Parameters**
   - o Initializes parameters controlling the evolutionary process, such as population_size, generations, similarity_threshold, elite_percentage, etc.
   - o Validates parameters like elite_percentage to ensure they fall within acceptable ranges.

9. **Early Stopping Configuration**
   - o Validates and sets up early stopping based on the provided metric and value.
   - o Ensures that necessary parameters (early_stopping_value) are provided when an early stopping metric is specified.

10. **Fitness and Caching**
    - o self.baseline_mse: Computes the mean squared error of a baseline model (predicting the mean of the target), serving as a reference point.
    - o self.expression_cache: Initializes a cache for storing evaluated expressions, with a maximum size to manage memory usage.

11. **Diversity and Mutation Setup**
    - o self.diversity_weight: Balances the importance of similarity versus fitness in selecting expressions.
    - o self.mutation_types: Lists the types of mutations ('replace', 'add', 'remove') applicable during evolution.

12. **Operator and Constant Definitions**
    - o Defines lists of unary and binary operators (e.g., sin, cos, Add, Mul, Div) that can be used in constructing expressions.
    - o self.constant_choices: Provides a set of constant values (integers, mathematical constants like $\pi$ and e, fractions, square roots) that can be incorporated into expressions.
    - o self.variable_and_constant_choices: Combines variables and constants, serving as the pool from which expression components are drawn.

13. **Lambdification Setup**
    - o self.lambdify_modules: Specifies the modules and functions to be used when converting SymPy expressions into callable functions via sympy.lambdify. This includes mapping SymPy functions to their NumPy equivalents for numerical computations.

14. **Sample Values and Unit Array**
    - o self.unit_array: An array of ones, possibly used as a placeholder or for normalization purposes.
    - o self.sample_value_arrays: Stores sample values for each variable, initialized with the unit array.

15. **Last Resort Mechanism**

- o self.last_resort: Indicates whether a last-resort strategy should be activated if the primary regression process fails.
- o self.last_resort_activated: Tracks whether the last-resort mechanism has been triggered.
16. **Logging Initialization**
    - o Logs the initialization parameters and configurations, providing a comprehensive overview of the regression setup.

**Method: get_expression_depth_cached(self, expr)**
This method retrieves the cached depth of expr if available; otherwise, computes and caches it.

**Method: get_expression_depth(self, expr)**
A public method to obtain the depth of an expression using caching.

**Method: Safe Mathematical Operations**
Ensures mathematical operations are robust against invalid inputs (e.g., division by zero, logarithm of non-positive numbers).

**Variable Relevance Assessment**
- o assess_variable_relevance(self, relevance_metric='correlation')
    - Evaluates the relevance of each input variable based on specified metrics (correlation, mutual information, or both).
- o Parameters:
    - relevance_metric (str): One of 'correlation', 'mutual_information', or 'both' [calculated with a weight of 0.5 each].
- o Behavior:
    - Computes Pearson correlation coefficients and/or mutual information scores between each variable and the target.

**Population Initialization and Management**
- initialize_population(self)
    - o Generates the initial population of symbolic expressions using a combination of random and user-provided expressions.
    - o Behavior:
        - Creates expressions of varying types (simple, unary, binary, complex) up to population_size.
        - Ensures inclusion of all variables if force_all_variables is enabled.
- generate_random_expression(self, depth=0, used_vars=None)
    - o Recursively generates a random symbolic expression respecting the maximum depth constraint.
    - o Parameters:
        - depth (int): Current depth of recursion.
        - used_vars (set): Set of variables already included in the expression.
    - o Behavior:
        - Chooses to generate a leaf node or apply unary/binary operators based on probabilistic decisions.
        - Mathematical Notation: E={v, if leaf or f(E1,E2,…) if operator applied}
- generate_leaf(self)
    - o Selects a random variable or constant to serve as a leaf node in an expression tree.
    - o **Mathematical Notation:** v∈{x1,x2,…,xn}∪{constants}
- prune_population(self)
    - o Filters the population to retain only valid and unique expressions, optionally enforcing variable inclusion.
    - o Behavior:

- Removes invalid expressions (e.g., those causing errors or exceeding constraints).
- Ensures population size does not exceed population_size by retaining top-performing expressions.
- Mathematical Notation: $P'=\text{Top}_{population\_size}(P)$. Where $P'$ is the pruned population.

**Expression Evaluation and Fitness Calculation**

- **evaluate_expression(self, expression, timeout=5)**
    - Evaluates the fitness of a symbolic expression by computing performance metrics against the target data.
    - Parameters:
        - expression (SymPy expression or str): The expression to evaluate.
        - timeout (float): Maximum time allowed for evaluation.
    - Returns: Tuple containing fitness score and various metrics (MSE, $R^2$, etc.).
    - Mathematical Notation: $\text{Fitness}=\text{Accuracy\_Score} \times \text{Complexity\_Penalty}$
- **cached_evaluate_expression(self, expression_str, timeout=5)**
    - Caches the evaluation results of expressions to avoid redundant computations.
    - Parameters:
        - expression_str (str): String representation of the expression.
        - timeout (float): Evaluation timeout.
    - Returns: Evaluation results from evaluate_expression.
- **is_valid_expression(self, expr)**
    - Checks whether an expression is valid by ensuring it can be evaluated without errors and does not produce undefined values.
    - Behavior:
        - Attempts to lambdify and evaluate the expression on sample data.
        - Mathematical Notation: $\text{Valid} \Leftrightarrow \text{Expression Evaluates Without Errors and No NaN/Inf}$

**Genetic Operators**

- **mutate_expression(self, expression)**
    - Applies a mutation operation (replace, add, or remove) to an expression to generate a new variant.
    - Parameters:
        - expression (SymPy expression): The expression to mutate.
    - Returns: Mutated expression.
    - Mathematical Notation: $E_{mutated}=\text{Mutate}(E)$
- **crossover_expressions(self, expr1, expr2)**
    - Combines two parent expressions by swapping random subtrees to produce offspring.
    - Parameters:
        - expr1, expr2 (SymPy expressions): Parent expressions.
    - Returns: New expression resulting from crossover.
    - Mathematical Notation: $E_{offspring}=\text{Crossover}(E_1, E_2)$
- **merge_expressions(self, expr1, expr2)**
    - Merges two expressions using a randomly selected operator (e.g., addition or multiplication).
    - Parameters:
        - expr1, expr2 (SymPy expressions): Expressions to merge.
    - Returns: Merged expression.
    - Mathematical Notation: $E_{merged}=f(E_1,E_2)$, where $f \in \{\text{Add, Mul, etc.}\}$

**Selection Mechanisms**

- **select_diverse_expression(self, population)**
    - Selects an expression from the population aiming to maximize diversity based on similarity scores.
    - Parameters:

- population (list): Current population with fitness scores.
  - Returns: Selected expression.
  - Mathematical Notation: Select E such that max(Diversity(E,P)) is maximized
- **select_similar_expression(self, target_expr, population)**
  - Selects an expression similar to target_expr based on the similarity threshold.
  - Parameters:
    - target_expr (SymPy expression): Reference expression.
    - population (list): Current population with fitness scores.
  - Returns: Similar expression if available; otherwise, random selection.

### Evolutionary Process
- **evolve(self)**
  - Executes the main evolutionary loop, iterating through generations to evolve the population towards optimal expressions.
  - Behavior:
    1. Initializes population.
    2. Iteratively evaluates fitness, selects elites, applies genetic operators, and updates population.
    3. Monitors for early stopping conditions and timeout.
    4. Returns the best-found expression.
  - Mathematical Notation: for each generation g=1 to G:{Evaluate Fitness($P_g$), Select Elites, Apply Mutation/Crossover, Form $P_{g+1}$

### Similarity and Diversity Management
- **calculate_similarity(self, expr1, expr2)**
  - Computes similarity between two expressions based on tree edit distance normalized by expression size.
  - Parameters:
    - expr1, expr2 (SymPy expressions): Expressions to compare.
  - Returns: Similarity score in [0, 1].
  - Mathematical Notation: Similarity(E1,E2)=1−(TreeEditDistance(E1,E2))/(max(Size(E1),Size(E2)))
- **tree_edit_distance(self, tree1_str, tree2_str, depth=0)**
  - Recursively calculates the tree edit distance between two symbolic expressions.
  - Parameters:
    - tree1_str, tree2_str (str): String representations of the expressions.
  - Returns: Normalized similarity score.
  - Mathematical Notation: TreeEditDistance(T1,T2)=∑(insertions, deletions, substitutions)
- **check_diversity(self)**
  - Assesses population diversity and introduces new random expressions if diversity falls below a threshold.
  - Behavior:
    - Computes diversity as the ratio of unique expressions.
    - Adds 20% new random expressions if diversity <0.5< 0.5<0.5.
  - Mathematical Notation: Diversity=|{Ei}||P|, if Diversity<0.5⇒Add 0.2|P| random expressions

### Utility and Helper Methods
- **get_all_subexpressions_cached(self, expr_str)**
  - Retrieves all subexpressions of a given expression, utilizing caching for efficiency.
  - Parameters:
    - expr_str (str): String representation of the expression.
  - Returns: List of subexpressions.
- **get_random_subexpression(self, expr)**

- o Selects a random subexpression from a given expression.
  - o Parameters:
    - ▪ expr (SymPy expression): The expression from which to extract a subexpression.
  - o Returns: Random subexpression.
- truncate_expression(self, expr)
  - o Reduces the depth of an expression by truncating its subtrees to adhere to max_depth.
  - o Parameters:
    - ▪ expr (SymPy expression): The expression to truncate.
  - o Returns: Truncated expression.
- insert_user_expression_subtree(self, expr)
  - o Inserts a user-provided expression subtree into a random location within another expression.
  - o Parameters:
    - ▪ expr (SymPy expression): The target expression for insertion.
  - o Returns: Modified expression with the user subtree.
- get_lambdified_function(self, expression)
  - o Converts a SymPy expression into a NumPy-compatible function using lambdify, with caching.
  - o Parameters:
    - ▪ expression (SymPy expression): The expression to lambdify.
  - o Returns: Callable function or None if conversion fails.
- update_similarity_threshold(self)
  - o Dynamically adjusts the similarity threshold based on population diversity.
  - o Behavior:
    - ▪ Sets similarity_threshold within [0.1, 0.9] inversely proportional to diversity.
  - o Mathematical Notation: $similarity\_threshold = \max(0.1, \min(0.9, initial\_threshold \times (1 - diversity)))$
- update_elite_percentage(self)
  - o Dynamically adjusts the elite percentage based on diversity and evolutionary progress.
  - o Behavior:
    - ▪ Increases elite_percentage up to a maximum as diversity decreases.
  - o Mathematical Notation: $elite\_percentage = \min(max\_elite, base\_elite + progress \times (1 - diversity) \times (max\_elite - base\_elite))$

### Cache Management
- clear_all_lru_caches()
  - o Clears all LRU caches used by cached functions to reset cached data.

### Additional Methods
- dynamic_complexity_penalty(self, complexity, generation, best_fitness)
  - o Computes a penalty factor based on expression complexity, generation progress, and fitness to balance exploration and exploitation.
  - o Mathematical Notation: $Penalty = e^{-\beta}$, where $\beta = base\_penalty \times fitness\_factor \times complexity\_scale \times diversity\_factor$
- find_similar_expressions(self, expression)
  - o Identifies expressions in the population that meet or exceed the similarity threshold relative to a given expression.
  - o Parameters:
    - ▪ expression (SymPy expression): Reference expression.
  - o Returns: List of similar expressions.
- determine_missing_variables_cause(self, missing_vars)
  - o Diagnoses potential reasons for missing variables in the best expression.
  - o Parameters:
    - ▪ missing_vars (set): Variables not present in the best expression.
  - o Returns: String summarizing possible causes.

The complete class of SPINEX is shown in the Appendix.

**3.0 Benchmarking of the proposed algorithm**

Benchmarking is a critical component in the evaluation of newly developed symbolic regression algorithms. In this study, the benchmarking was conducted by utilizing a number of standardized datasets and problem instances that enable researchers to assess the performance, accuracy, and efficiency of different symbolic regression methodologies. Among the various benchmarking suites, the Feynman, Keijzer, Korns, Livermore, Nguyen, and R rationals problem sets are used, given their diverse and challenging problem sets. Each of these databases is described in this section (with additional details available from the cited sources).

*3.1. Feynman problem set*

The Feynman database comprises a collection of physical equations derived from the Feynman Lectures on Physics. It is designed to test the capability of symbolic regression algorithms to rediscover fundamental physical laws from empirical data. The database includes a wide range of equations covering mechanics, electromagnetism, thermodynamics, and quantum physics. Each equation is accompanied by data generated to simulate real-world measurement uncertainties. This database seems to appear first in Udrescu and Tegmark [12] (who used generic input values for each function, ranging between 1 and 5), and then in Matsubara et al. [27] (who further refined the functions and provided actual/practical input values for each function and corresponding parameters). Overall, this database can be further grouped into three sets, often noted as Easy, Medium, and Hard – to describe each set's difficulty level. In total, there are 120 functions in this problem set and these are not provided herein for brevity but can still be accessed via the above-cited sources.

*3.2. Keijzer problem set*

Developed by Keijzer [28], the Keijzer problem set is a compilation of randomly generated benchmark problems specifically tailored for symbolic regression. It consists of a diverse set of mathematical functions varying in complexity and dimensionality, including polynomials and power functions. There are 11 functions in this set, as seen in Table 1.

Table 1 Keijzer problem set

| No. | Function |
|---|---|
| 1 | $f(x) = 0.3 \cdot x \cdot sin(2 \cdot 3.14 \cdot x)$ |
| 2 | $f(x) = x^3 \cdot e^{-x} \cdot \cos(x) \cdot \sin(x) \cdot (sin^2(x) \cdot cos(x) - 1)$ |
| 3 | $f(x) = \frac{30 \cdot x \cdot z}{(x-10) \cdot y^2}$ |
| 4 | $f(x) = \log(x)$ |
| 5 | $f(x) = \sqrt{x}$ |
| 6 | $f(x,y) = x^y$ |
| 7 | $f(x,y) = x \cdot y + \sin((x-1)(y-1))$ |
| 8 | $f(x,y) = x^4 - x^3 + \frac{y^2}{2} - y$ |
| 9 | $f(x,y) = 6 \cdot \sin(x) \cdot \cos(y)$ |
| 10 | $f(x,y) = \frac{8}{2+x^2+y^2}$ |
| 11 | $f(x,y) = \frac{x^3}{5} + \frac{y^3}{2} - y - x$ |

*Range of inputs: x (0, 10) expect in No. 3 (1, 100), No. 4 (0, 100) and No. 7 (1, 10). y (0, 10) except in No. 3 and 7 (1, 10) and (1, 5), respectively. z (0.10)*

### 3.3. Korns problem set

The Korns database, named after its creator [29], offers a suite of 15 problems that emphasize real-world applicability and complexity. This set includes equations that contain no more than three grammar nodes deep, with all test problems reference no more than five input features. The Korns database is distinguished by its inclusion of multi-variable and non-linear equations. Table 2 lists such functions.

Table 2 Korns problem set

| No. | Function | Input range |
|---|---|---|
| 1 | $y = 1.57 + 24.3 \cdot X_3$ | $X_3 \in [0,10]$ |
| 2 | $y = 0.23 + 14.2 \cdot \frac{(X_3+X_1)}{3.0 \cdot X_4}$ | $X_1$ and $X_3 \in [0,10]$, $X_4 \in [1,10]$ |
| 3 | $y = 0.23 + 14.2 \cdot \frac{(X_3+X_1)}{3.0 \cdot X_4}$ | $X_0$-$X_4 \in [1,10]$ |
| 4 | $y = 3.0 + 2.13 \cdot \log(X_4)$ | $X_4 \in [1,10]$ |
| 5 | $y = 1.3 + 0.13 \cdot \sqrt{X_0}$ | $X_0 \in [0,10]$ |
| 6 | $y = 213.80940889 - 213.80940889 \cdot e^{-0.54723748542 \cdot X_0}$ | $X_0 \in [0,10]$ |
| 7 | $y = 6.87 + 11 \cdot \sqrt{7.23 \cdot X_0 \cdot X_3 \cdot X_4}$ | $X_0$, $X_3$ and $X_4 \in [0,10]$ |
| 8 | $y = \left(\frac{\sqrt{X_0}}{\log(X_1)}\right) \cdot \left(\frac{e^{X_2}}{X_3^2}\right)$ | $X_0$, $X_1$ and $X_3 \in [1,10]$, $X_2 \in [0,10]$ |
| 9 | $y = 0.81 + 24.3 \cdot \frac{(2.0 \cdot X_1 + 3.0 \cdot X_2^2)}{4.0 \cdot X_3^3 + 5.0 \cdot X_4^4}$ | $X_1$ and $X_2 \in [0,10]$, $X_3$ and $X_4 \in [1,10]$, |
| 10 | $y = 6.87 + 11 \cdot \cos(7.23 \cdot X_0^3)$ | $X_0 \in [0,10]$ |
| 11 | $y = 2.0 - 2.1 \cdot (\cos(9.8 \cdot X_0) \cdot \sin(1.3 \cdot X_4))$ | $X_0$ and $X_4 \in [0,10]$ |
| 12 | $y = 32.0 - 3.0 \cdot \left(\frac{\tan(X_0)}{\tan(X_1)} \cdot \frac{\tan(X_2)}{\tan(X_3)}\right)$ | $X_0, X_1, X_2, X_3 \in [0,\pi]$ |
| 13 | $y = 22.0 + 4.2 \cdot \left((\cos(X_0) - \tan(X_1)) \cdot \frac{\tanh(X_2)}{\sin(X_3)}\right)$ | $X_0, X_1 \in [0,\pi]$, $X_2 \in [-5,5]$, $X_3 \in [0.1,\pi]$ |
| 14 | $y = 12.0 - 6.0 \cdot \left(\frac{\tan(X_0)}{e^{X_1}} \cdot (\log(X_2) - \tan(X_3))\right)$ | $X_0 \in [0,\pi]$, $X_1 \in [0,10]$, $X_2 \in [1,10]$, $X_3 \in [0,\pi]$ |

### 3.4. Livermore problem set

The Livermore database originates from the Lawrence Livermore National Laboratory and consists of a set of benchmark functions designed to evaluate symbolic regression techniques [30]. The Livermore database is particularly useful for assessing the performance of SR methods in handling large-scale and computationally intensive tasks, thereby providing insights into their practical applicability in high-demand scenarios. There are 22 functions in this set, which can be accessed from [30].

Table 3 Nguyen problem set

| No. | Function | Input range |
|---|---|---|
| 1 | $F_1(x) = \frac{1}{3} + x + \sin(x^2)$ | [-10,10] |
| 2 | $F_2(x) = \sin(x^2) \cdot \cos(x) - 2$ | [-1,1] |
| 3 | $F_3(x) = \sin(x^3) \cdot \cos(x^2) - 1$ | [-1,1] |
| 4 | $F_4(x) = \log(x + 1) + \log(x^2 + 1) + \log(x)$ | [0,2] |

| 5 | $F_5(x,y) = x^4 - x^3 + x^2 - y$ | [0,1] |
| 6 | $F_6(x) = 4x^4 + 3x^3 + 2x^2 + x$ | [-1,1] |
| 7 | $F_7(x) = \sinh(x)$ | [-1,1] |
| 8 | $F_8(x) = \cosh(x)$ | [-1,1] |
| 9 | $F_9(x) = x^9 + x^8 + x^7 + x^6 + x^5 + x^4 + x^3 + x^2 + x$ | [-1,1] |
| 10 | $F_{10}(x,y) = 6 \cdot \sin(x) \cdot \cos(y)$ | [0,1] |
| 11 | $F_{11}(x,y) = \frac{x^2 \cdot x^2}{x+y}$ | [-1,1] |
| 12 | $F_{12}(x,y) = \frac{x^5}{y^3}$ | [-1,1] |
| 13 | $F_{13}(y) = y^{\frac{1}{3}}$ | [0,4] |
| 14 | $F_{14}(x) = x^3 + x^2 + x + \sin(x) + \sin(x^2)$ | [-1,1] |
| 15 | $F_{15}(x) = x^{\frac{1}{5}}$ | [0,4] |
| 16 | $F_{16}(x) = x^{\frac{2}{5}}$ | [0,4] |
| 17 | $F_{17}(x,y) = 4 \cdot \sin(x) \cdot \cos(y)$ | [0,1] |
| 18 | $F_{18}(x) = \sin(x^2) \cdot \cos(x) - 5$ | [-1,1] |
| 19 | $F_{19}(x) = x^5 + x^4 + x^2 + x$ | [-1,1] |
| 20 | $F_{20}(x) = e^{-x^2}$ | [-1,1] |
| 21 | $F_{21}(x) = x^8 + x^7 + x^6 + x^5 + x^4 + x^3 + x^2 + x$ | [-1,1] |
| 22 | $F_{22}(x) = e^{-0.5 \cdot x^2}$ | [-1,1] |

### 3.5. Nguyen problem set

The Nguyen [16] problem set is a widely recognized benchmark suite in the symbolic regression community, comprising a series of mathematical expressions ranging from simple to highly complex. These functions are typically polynomial and transcendental in nature, with varying degrees of noise and dimensionality. There are 12 functions and these can be seen in Table 4.

Table 4 Nguyen problem set

| No. | Function |
|---|---|
| 1 | $F_1(x) = x^3 + x^2 + x$ |
| 2 | $F_2(x) = x^4 + x^3 + x^2 + x$ |
| 3 | $F_3(x) = x^5 + x^4 + x^3 + x^2 + x$ |
| 4 | $F_4(x) = x^6 + x^5 + x^4 + x^3 + x^2 + x$ |
| 5 | $F_5(x) = \sin(x^2) \cdot \cos(x) - 1$ |
| 6 | $F_6(x) = \sin(x) + \sin(x + x^2)$ |
| 7 | $F_7(x) = \log(x+1) + \log(x^2+1)$ |
| 8 | $F_8(x) = \sqrt{x}$ |
| 9 | $F_9(x,y) = \sin(x) + \sin(y^2)$ |
| 10 | $F_{10}(x,y) = 2 \cdot \sin(x) \cdot \cos(y)$ |
| 11 | $F_{11}(x,y) = x^y$ |
| 12 | $F_{12}(x,y) = \frac{1}{1+x^{-4}} + \frac{1}{1+y^{-4}}$ |

*Range of inputs: x and y (0, 10) expect in No. 8 (0, 100), No. 12 (1, 10)*

### 3.6. R rationals problem set

The R rationals set was introduced by Krawiec and Pawlak [31] and includes three functions (with two different input ranges for each function). Table 5 shows these functions.

Table 5 R rationals problem set

| No. | Function |
|---|---|
| 1 | $F_1(x) = \frac{(x+1)^3}{x^2-x+1}$ |
| 2 | $F_2(x) = \frac{x^5-3x^3+1}{x^2+1}$ |
| 3 | $F_3(x) = \frac{x^6+x^5}{x^4+x^3+x^2+x+1}$ |

*Range of inputs: x (-1, 1) and (-10, 10)*

## 4.0 Description of benchmarking experiments and metrics

In this evaluation, a comprehensive set of metrics was employed to assess the performance of the proposed algorithm in arriving at target expressions. The metrics encompass both quantitative measures of predictive accuracy and qualitative assessments of model structure alignment with the actual functions. The primary metrics utilized include Coefficient of Determination ($R^2$), and a Composite Metric (CM) that integrates multiple performance aspects. Additionally, comparisons between actual and predicted variables and operations provide insights into the structural fidelity of the models.

The aforenoted CM was devised by integrating $R^2$, variable set match, and operation set match. This metric is formulated as:

$$CM = w_{R^2} \cdot R^2 + w_{var} \cdot V + w_{op} \cdot O$$

where:

- $V$ is a binary indicator representing whether the set of variables in the predicted model matches the actual model. The Variable Set Match assesses the extent to which the variables utilized in the predicted symbolic expression align with those in the actual expression. It is a binary metric where:

    o $Variable\ Set\ Match = \{1, if\ V_{actual} = V_{predicted}, and\ 0\ otherwise\}$

- $O$ is a binary indicator representing whether the set of operations (e.g., addition, sine) in the predicted model matches those in the actual model. The Operation Set Match evaluates the correspondence between the mathematical operations (e.g., addition, sine, logarithm) employed in the actual and predicted expressions, and is defined as:

    o $Operation\ Set\ Match = \{1, if\ O_{actual} = O_{predicted}, and\ 0\ otherwise\}$

- $w_{R^2} = 0.5, w_{var} = 0.25, w_{op} = 0.25$ are the respective weights assigned to each component based on their importance.

In addition, while a number of symbolic regression algorithms exist, we opt to compare our analysis against PySR, as it has been noted as one of the leading symbolic regression algorithms, as seen in recent competitions and publications. Another reason to utilize PySR is the fact that it also builds on the use of genetic evolutionary (vs. deep learning, as seen in [27]). PySR autonomously generates and optimizes mathematical expressions that best fit the provided data by balancing both accuracy and model simplicity. PySR also employs multi-objective optimization to simultaneously minimize both the models' loss and complexity. By utilizing Pareto front analysis, PySR identifies a set of non-dominated solutions where no single model is unequivocally superior across all objectives. This approach ensures that the resulting models offer an optimal balance between accuracy and simplicity. PySR was used with its default settings, as seen in [44](just like how SPINEX was also used in its default settings). It is worth noting that we attempted to use other algorithms (such as gplearn), however, such algorithms struggled on a number of problem sets or yielded poor results that prevented us from conducting a comprehensive examination. PySR provided much more consistent results, so this algorithm is selected for primary comparison.

This benchmarking analysis involves a set of 182 mathematical benchmarking functions. This analysis was run and evaluated in a Python 3.10.5 environment using an Intel(R) Core(TM) i7-9700F CPU @ 3.00GHz and a RAM of 32.0GB.

*4.1 Benchmarking on problem sets*

In total, 182 benchmarking functions were used to examine the performance of the proposed algorithm. This section discusses the outcome of this analysis. It is worth noting that the SPINEX was used with the following settings: population_size (50 and 100), number_of_generations (100 and 1000), and force_all_variables (True, False). All other settings were kept at their default values (see Appendix). The primary goal of the selected settings was to examine the influence of population size and number of generations with the inclusion of all variables (or not) on the observed response of the derived expressions. It is worth noting that SPINEX was allowed to terminate early once arriving at an expression with a maximum fitness value (i.e., $R^2 = 1.0$).

The benchmarking analysis presented in Table 6 and Fig. 1 compare the performance of two symbolic regression models, SPINEX and PySR, across the above-selected problem sets. The evaluation metrics include $R^2$ values and the rates of Exact Variable Set Match (Exact VM) and Exact Operation Set Match (Exact OP). Overall, one can clearly see that PySR seems to lead on the accuracy ($R^2$) front over SPINEX in most of the problems (while SPINEX outperforms PySR in two instances). However, SPINEX demonstrates a notable advantage in achieving higher Exact VM percentages, indicating that SPINEX significantly outperforms PySR across most problem sets when forcing all variables, achieving up to 100% in certain configurations. The Exact OP metric also favors SPINEX, particularly under conditions where all variables are forced, suggesting a higher fidelity in matching both variable and operation sets. The performance of both algorithms is also noted to improve, generally, with the use of larger population and generations. This holistic analysis showcases the merit of utilizing SPINEX on well-defined problem sets (especially in those where all variables are expected to be present in the final expression).

Table 6 Outcome of benchmarking analysis

| | SPINEX | | PySR | SPINEX | | PySR | SPINEX | | PySR |
|---|---|---|---|---|---|---|---|---|---|
| R² | Count (force_all_variables=False) | Count (force_all_variables=True) | | Count (force_all_variables=False) | Count (force_all_variables=True) | | Count (force_all_variables=False) | Count (force_all_variables=True) | |
| | Feynman (Easy problem set with pop = 50, gen = 100) | | | Feynman (Medium problem set with pop = 50, gen = 100) | | | Feynman (Hard problem set with pop = 50, gen = 100) | | |
| R² < 0 | 36.00% | 44.00% | 41.67% | 48.57% | 45.71% | 40.00% | 12.50% | 10.42% | 6.25% |
| 0.0-0.1 | 0.00% | 0.00% | 0.00% | 2.86% | 0.00% | 0.00% | 0.00% | 2.08% | 0.00% |
| 0.1-0.2 | 0.00% | 0.00% | 0.00% | 0.00% | 0.00% | 0.00% | 0.00% | 2.08% | 0.00% |
| 0.2-0.3 | 0.00% | 0.00% | 0.00% | 0.00% | 0.00% | 0.00% | 4.17% | 2.08% | 0.00% |
| 0.3-0.4 | 0.00% | 0.00% | 0.00% | 0.00% | 0.00% | 0.00% | 2.08% | 2.08% | 0.00% |
| 0.4-0.5 | 4.00% | 0.00% | 0.00% | 2.86% | 2.86% | 0.00% | 4.17% | 2.08% | 0.00% |
| 0.5-0.6 | 4.00% | 4.00% | 0.00% | 0.00% | 0.00% | 0.00% | 2.08% | 0.00% | 2.44% |
| 0.6-0.7 | 4.00% | 8.00% | 0.00% | 0.00% | 0.00% | 0.00% | 2.08% | 4.17% | 0.00% |
| 0.7-0.8 | 0.00% | 0.00% | 0.00% | 2.86% | 0.00% | 0.00% | 8.33% | 8.33% | 0.00% |
| 0.8-0.9 | 0.00% | 8.00% | 0.00% | 5.71% | 8.57% | 0.00% | 6.25% | 14.58% | 0.00% |
| 0.9-1.0 | 52.00% | 36.00% | **58.33%** | 37.14% | 42.86% | **57.14%** | 58.33% | 52.08% | **91.67%** |
| force_all_variables | Exact VM (%) | Exact OP (%) | - | Exact VM (%) | Exact OP (%) | - | Exact VM (%) | Exact OP (%) | - |
| FALSE | 28.00% | 16.00% | Exact VM (%) = 20.83 | 11.43% | 5.71% | Exact VM (%) = 17.14 | 4.17% | 4.17% | Exact VM (%) = **4.17** |
| TRUE | **68.00%** | 16.00% | Exact OP (%) = 25.00 | **71.43%** | 5.71% | Exact OP (%) = 28.57 | **85.42%** | 4.17% | Exact OP (%) = 39.58 |
| | Feynman (Easy problem set with pop = 100, gen = 1000) | | | Feynman (Medium problem set with pop = 100, gen = 1000) | | | Feynman (Hard problem set with pop = 100, gen = 1000) | | |
| R² < 0 | 28.00% | 24.00% | 45.83% | 29.41% | 25.71% | 40.00% | 8.51% | 9.76% | 4.17% |
| 0.0-0.1 | 0.00% | 0.00% | 0.00% | 0.00% | 2.86% | 0.00% | 0.00% | 0.00% | 0.00% |
| 0.1-0.2 | 0.00% | 0.00% | 0.00% | 0.00% | 0.00% | 0.00% | 0.00% | 0.00% | 0.00% |
| 0.2-0.3 | 0.00% | 0.00% | 0.00% | 0.00% | 0.00% | 0.00% | 2.13% | 0.00% | 0.00% |
| 0.3-0.4 | 0.00% | 0.00% | 0.00% | 0.00% | 2.86% | 0.00% | 0.00% | 0.00% | 0.00% |
| 0.4-0.5 | 0.00% | 8.00% | 0.00% | 0.00% | 0.00% | 0.00% | 0.00% | 0.00% | 0.00% |
| 0.5-0.6 | 0.00% | 0.00% | 0.00% | 0.00% | 0.00% | 0.00% | 0.00% | 2.44% | 2.44% |
| 0.6-0.7 | 0.00% | 0.00% | 0.00% | 2.94% | 0.00% | 0.00% | 2.13% | 0.00% | 0.00% |
| 0.7-0.8 | 0.00% | 0.00% | 0.00% | 5.88% | 5.71% | 0.00% | 0.00% | 0.00% | 0.00% |
| 0.8-0.9 | 0.00% | 0.00% | 0.00% | 2.94% | 0.00% | 0.00% | 10.64% | 4.88% | 0.00% |
| 0.9-1.0 | **72.00%** | 68.00% | 54.17% | 58.82% | **62.86%** | 57.14% | 76.60% | 82.93% | **93.75%** |
| force_all_variables | Exact VM (%) | Exact OP (%) | - | Exact VM (%) | Exact OP (%) | - | Exact VM (%) | Exact OP (%) | - |
| FALSE | 32.00% | 12.00% | Exact VM (%) = 16.67 | 14.71% | 2.94% | Exact VM (%) = 14.29 | 6.38% | 6.38% | Exact VM (%) = 2.28 |
| TRUE | **76.00%** | 32.00% | Exact OP (%) = 29.17 | **77.14%** | 5.71% | Exact OP (%) = 20.00 | **87.80%** | 2.44% | Exact OP (%) = 25.00 |
| | Keijzer problem set (with pop = 50, gen = 100) | | | Korns problem set (with pop = 50, gen = 100) | | | Livermore problem set (with pop = 50, gen = 100) | | |
| R² < 0 | 0.00% | 20.00% | 0.00% | 16.67% | 16.67% | 8.33% | 20.00% | 13.33% | 0.00% |
| 0.0-0.1 | 10.00% | 0.00% | 0.00% | 8.33% | 8.33% | 1667% | 0.00% | 0.00% | 0.00% |
| 0.1-0.2 | 10.00% | 0.00% | 0.00% | 0.00% | 0.00% | 8.33% | 0.00% | 0.00% | 0.00% |
| 0.2-0.3 | 0.00% | 0.00% | 0.00% | 8.33% | 0.00% | 0.00% | 0.00% | 0.00% | 0.00% |
| 0.3-0.4 | 0.00% | 0.00% | 0.00% | 0.00% | 0.00% | 0.00% | 0.00% | 0.00% | 0.00% |
| 0.4-0.5 | 0.00% | 0.00% | 0.00% | 0.00% | 0.00% | 0.00% | 0.00% | 0.00% | 0.00% |
| 0.5-0.6 | 0.00% | 0.00% | 10.00% | 0.00% | 0.00% | 0.00% | 0.00% | 0.00% | 0.00% |
| 0.6-0.7 | 0.00% | 10.00% | 0.00% | 8.33% | 8.33% | 0.00% | 0.00% | 6.67% | 0.00% |
| 0.7-0.8 | 10.00% | 10.00% | 0.00% | 8.33% | 8.33% | 0.00% | 0.00% | 6.67% | 0.00% |
| 0.8-0.9 | 10.00% | 0.00% | 0.00% | 8.33% | 16.67% | 0.00% | 13.33% | 0.00% | 0.00% |
| 0.9-1.0 | 60.00% | 60.00% | **90.00%** | 41.67% | 41.67% | **66.67%** | 66.67% | 73.33% | **100.00%** |
| force_all_variables | Exact VM (%) | Exact OP (%) | - | Exact VM (%) | Exact OP (%) | - | Exact VM (%) | Exact OP (%) | - |
| FALSE | 70.00% | 20.00% | Exact VM (%) = 70.00 | 50.00% | 16.67% | Exact VM (%) = 41.67 | **86.67%** | 0.00% | Exact VM (%) = 85.71 |
| TRUE | **100.00%** | 10.00% | Exact OP (%) = 40.00 | **91.67%** | 0.00% | Exact OP (%) = 41.67 | **86.67%** | 0.00% | Exact OP (%) = 28.57 |
| | Keijzer problem set (pop = 100, gen = 1000) | | | Korns problem set (pop = 100, gen = 1000) | | | Livermore problem set (pop = 100, gen = 1000) | | |
| R² < 0 | 0.00% | 0.00% | 0.00% | 16.67% | 16.67% | 8.33% | 7.14% | 7.14% | 0.00% |
| 0.0-0.1 | 0.00% | 0.00% | 0.00% | 8.33% | 8.33% | 8.33% | 0.00% | 0.00% | 0.00% |
| 0.1-0.2 | 10.00% | 10.00% | 0.00% | 0.00% | 0.00% | 0.00% | 0.00% | 0.00% | 0.00% |
| 0.2-0.3 | 0.00% | 0.00% | 0.00% | 8.33% | 0.00% | 0.00% | 0.00% | 0.00% | 0.00% |
| 0.3-0.4 | 0.00% | 0.00% | 0.00% | 0.00% | 0.00% | 8.33% | 0.00% | 0.00% | 0.00% |
| 0.4-0.5 | 0.00% | 0.00% | 0.00% | 0.00% | 0.00% | 0.00% | 0.00% | 0.00% | 0.00% |
| 0.5-0.6 | 10.00% | 0.00% | 10.00% | 0.00% | 0.00% | 0.00% | 0.00% | 0.00% | 0.00% |
| 0.6-0.7 | 0.00% | 20.00% | 0.00% | 8.33% | 8.33% | 0.00% | 7.14% | 0.00% | 0.00% |
| 0.7-0.8 | 0.00% | 0.00% | 0.00% | 8.33% | 8.33% | 0.00% | 7.14% | 7.14% | 0.00% |
| 0.8-0.9 | 20.00% | 0.00% | 0.00% | 8.33% | 16.67% | 0.00% | 0.00% | 0.00% | 0.00% |
| 0.9-1.0 | 60.00% | 70.00% | **90.00%** | 41.67% | 41.67% | **75.00%** | 78.57% | 85.71% | **100.00%** |
| force_all_variables | Exact VM (%) | Exact OP (%) | - | Exact VM (%) | Exact OP (%) | - | Exact VM (%) | Exact OP (%) | - |
| FALSE | 80.00% | 10.00% | Exact VM (%) = 70.00 | 50.00% | 16.67% | Exact VM (%) = 41.67 | **100.00%** | 7.14% | Exact VM (%) = 95.24 |
| TRUE | **100.00%**8.33% | 20.00% | Exact OP (%) = 70.00 | **91.67%** | 0.00% | Exact OP (%) = 25.00 | **100.00%** | 14.29% | Exact OP (%) = 28.57 |
| | Nguyen problem set (with pop = 50, gen = 100) | | | R rationals problem set (with pop = 50, gen = 100) | | | | | |
| R² < 0 | 0.00% | 8.33% | 0.00% | 0.00% | 0.00% | 0.00% | | | |
| 0.0-0.1 | 8.33% | 0.00% | 0.00% | 0.00% | 0.00% | 0.00% | | | |

| | | | | | | |
|---|---|---|---|---|---|---|
| 0.1-0.2 | 0.00% | 0.00% | 0.00% | 0.00% | 0.00% | 0.00% |
| 0.2-0.3 | 0.00% | 0.00% | 0.00% | 0.00% | 0.00% | 0.00% |
| 0.3-0.4 | 0.00% | 0.00% | 0.00% | 0.00% | 0.00% | 0.00% |
| 0.4-0.5 | 0.00% | 0.00% | 0.00% | 0.00% | 0.00% | 0.00% |
| 0.5-0.6 | 16.67% | 8.33% | 8.33% | 0.00% | 0.00% | 0.00% |
| 0.6-0.7 | 0.00% | 0.00% | 0.00% | 16.67% | 0.00% | 0.00% |
| 0.7-0.8 | 0.00% | 0.00% | 0.00% | 0.00% | 16.67% | 0.00% |
| 0.8-0.9 | 8.33% | 0.00% | 0.00% | 0.00% | 0.00% | 0.00% |
| 0.9-1.0 | 66.67% | 83.33% | **91.67%** | 83.33% | 83.33% | **100.00%** |
| force_all_variables | Exact VM (%) | Exact OP (%) | - | Exact VM (%) | Exact OP (%) | - |
| FALSE | 83.33% | 16.67% | Exact VM (%) = **91.67** | **100.00%** | 0.00% | Exact VM (%) = **100.00** |
| TRUE | **100.00%** | 16.67% | Exact OP (%) = 50.00 | **100.00%** | 33.33% | Exact OP (%) = 16.67 |
| Nguyen problem set (pop = 100, gen = 1000) | | | R rationals problem set (pop = 100, gen = 1000) | | | |
| $R^2 < 0$ | 0.00% | 8.33% | 0.00% | 0.00% | 0.00% | 0.00% |
| 0.0-0.1 | 8.33% | 0.00% | 0.00% | 0.00% | 0.00% | 0.00% |
| 0.1-0.2 | 0.00% | 0.00% | 0.00% | 0.00% | 0.00% | 0.00% |
| 0.2-0.3 | 0.00% | 0.00% | 0.00% | 0.00% | 0.00% | 0.00% |
| 0.3-0.4 | 0.00% | 0.00% | 0.00% | 0.00% | 0.00% | 0.00% |
| 0.4-0.5 | 0.00% | 0.00% | 0.00% | 0.00% | 0.00% | 0.00% |
| 0.5-0.6 | 16.67% | 8.33% | 0.00% | 0.00% | 0.00% | 0.00% |
| 0.6-0.7 | 0.00% | 0.00% | 0.00% | 16.67% | 0.00% | 0.00% |
| 0.7-0.8 | 0.00% | 0.00% | 0.00% | 0.00% | 16.67% | 0.00% |
| 0.8-0.9 | 8.33% | 0.00% | 0.00% | 0.00% | 0.00% | 0.00% |
| 0.9-1.0 | 66.67% | 83.33% | **100.00%** | 83.33% | 83.33% | **100.00%** |
| force_all_variables | Exact VM (%) | Exact OP (%) | - | Exact VM (%) | Exact OP (%) | - |
| FALSE | **83.33%** | 16.67% | Exact VM (%) = 75.00 | **100.00%** | 16.67% | Exact VM (%) = **100.00** |
| TRUE | **100.00%** | 16.67% | Exact OP (%) = 50.00 | **100.00%** | 16.67% | Exact OP (%) = 00.00 |

VM and VO are Exact Variable Set Match (%) and Exact Operation Set Match (%)

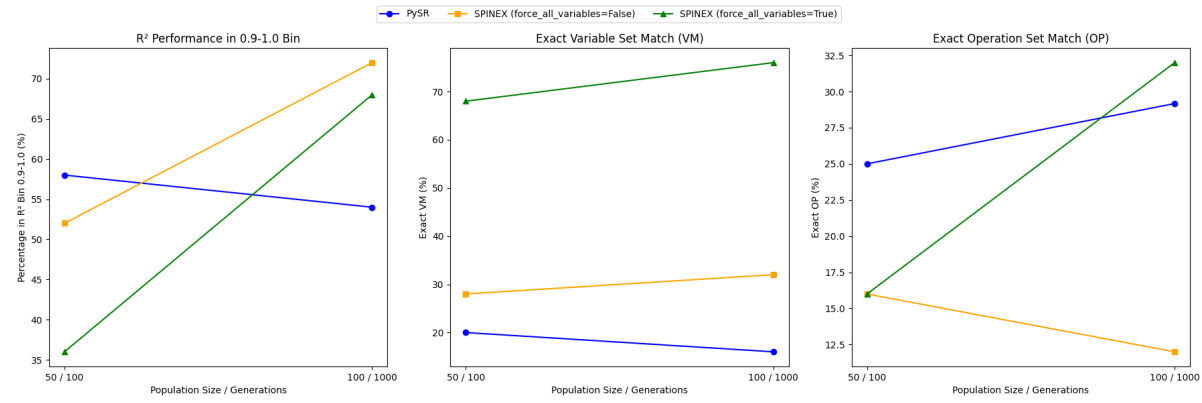
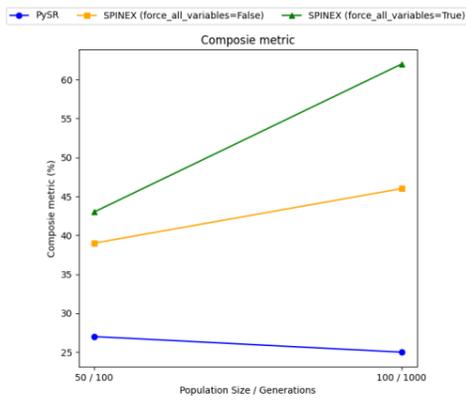

Feynman (Easy problem set)

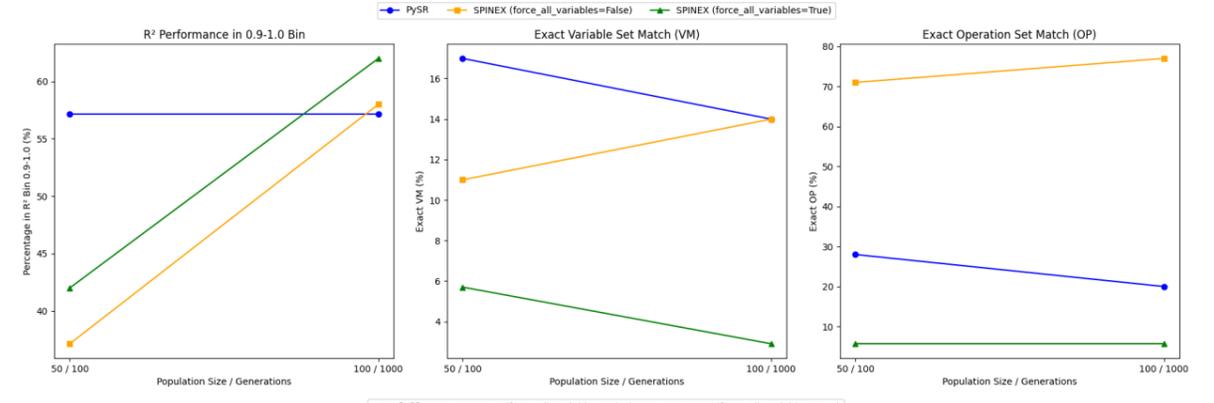
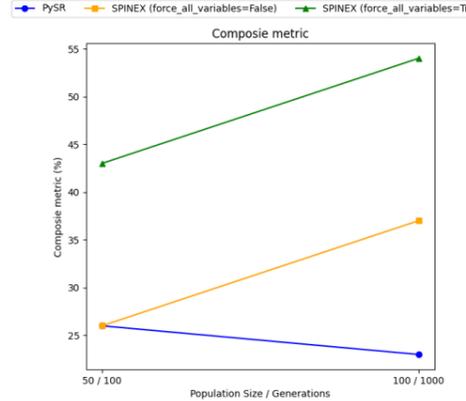

Feynman (Medium problem set)

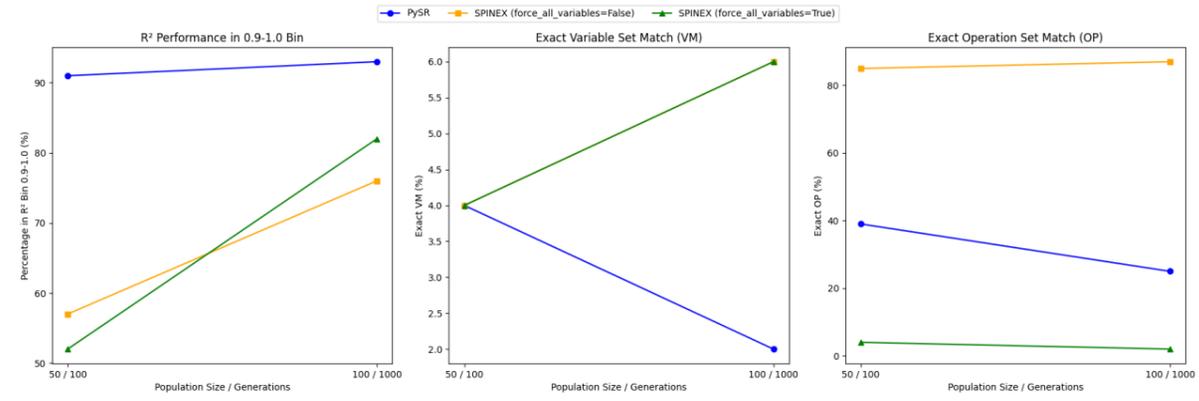
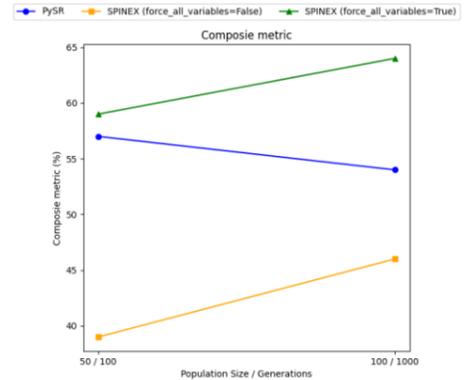

Feynman (Hard problem set)

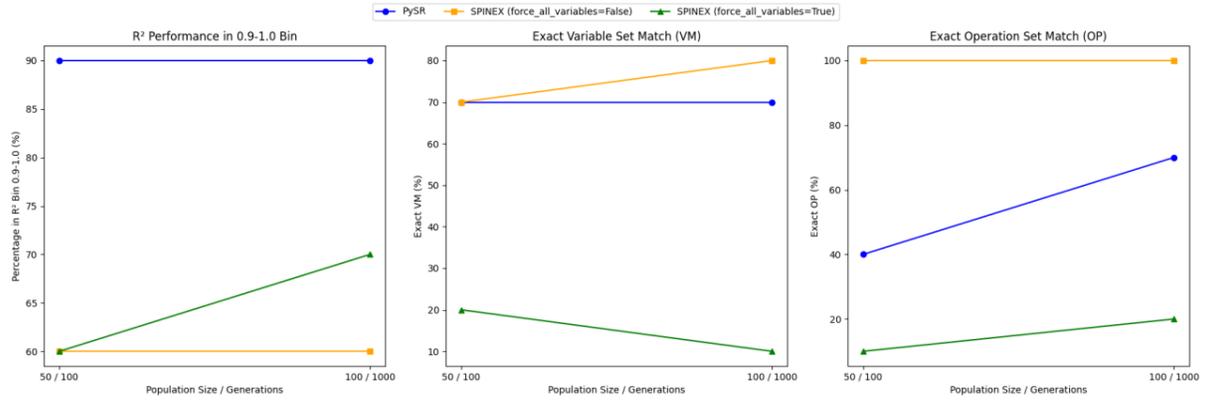
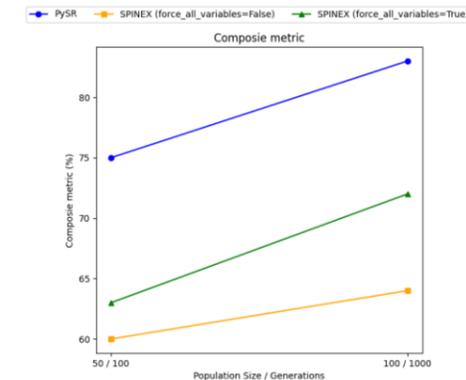

Keijzer problem set

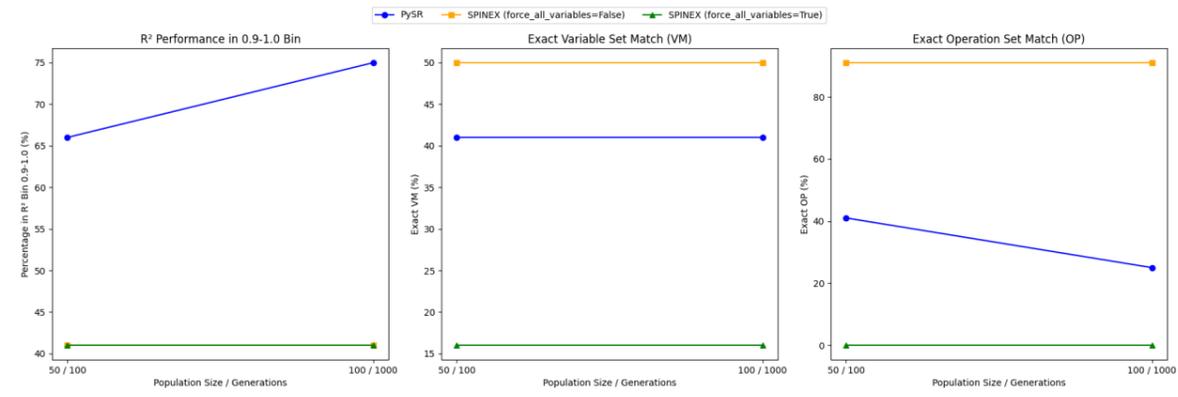
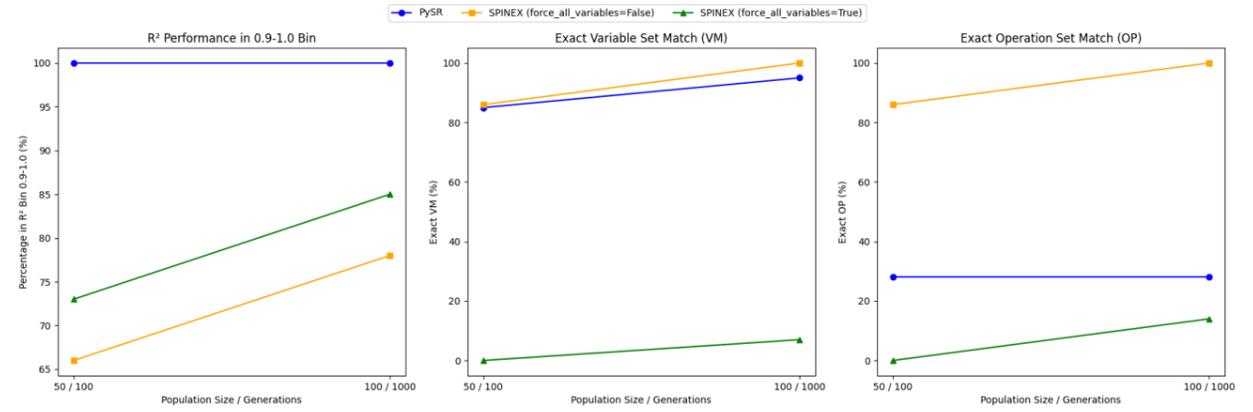
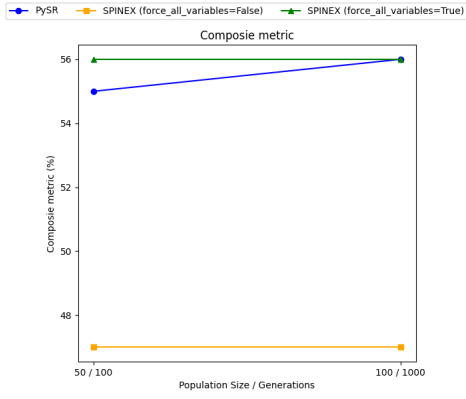
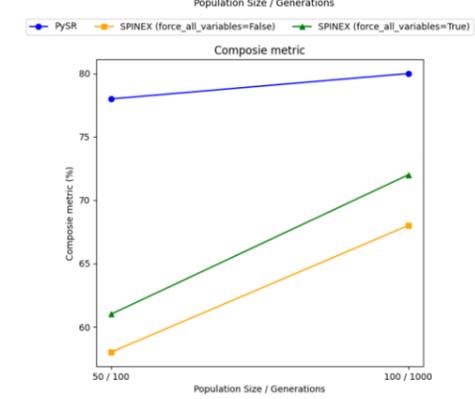

Korns problem set

Livermore problem set

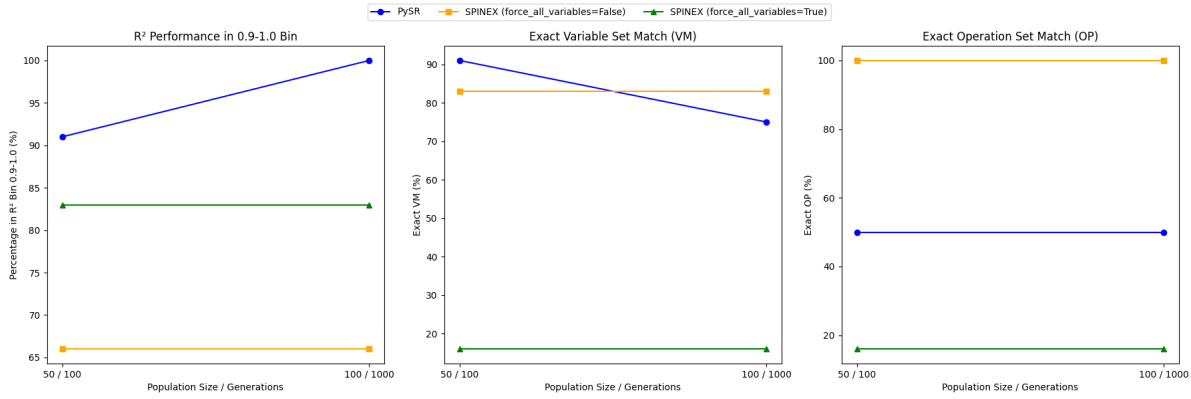
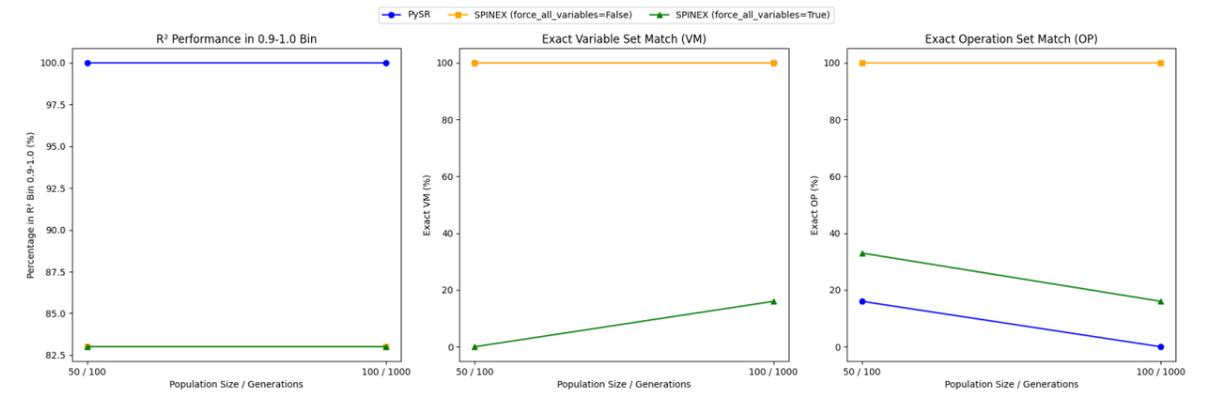
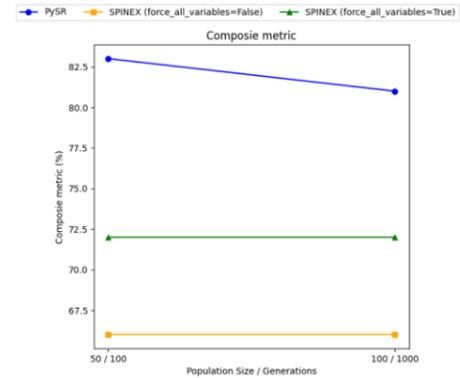
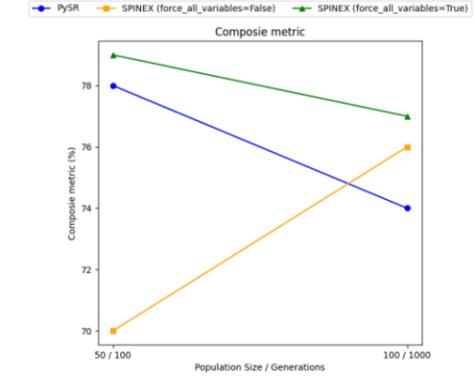

Nguyen problem set

R rationals problem set

Fig. 1 Further insights into the influence of population size and generation number on the outcome of this analysis (Note that in some of the problem sets, the composite results associated with PySR may be higher than SPINEX as this metric relies on the accuracy (which has a weight of 50% in this metric). Such results are to be examined by cross-checking the VM and OP metrics).

*4.2 Explainability analysis*

To showcase the explainability capability of SPINEX, an example is provided herein. In this example, SPINEX_SymbolicRegression is applied to a single-variable dataset generated by the quadratic function $y=2X^2+3X+5$. The dataset comprises 100 points evenly spaced between 0 and 10. The algorithm aims to discover the underlying mathematical relationship by evolving expressions through genetic operations over 50 generations with a population size of 50. The Advanced Explanation section provides a detailed breakdown of the evolved expressions. The first term, X, exhibits a negative fitness of -0.1698 and a high Mean Squared Error (MSE) contribution of 11070.5892, indicating poor performance in approximating the target function. It shows a strong correlation (0.9752) and perfect Spearman correlation (1.0000) with the actual data but suffers from low similarity measures and a negative $R^2$ value, suggesting it does not capture the underlying quadratic relationship. The second term, X+10X, also has a negative fitness (-0.0812) and a lower MSE (9530.5219) compared to the first term. While slightly more complex, it still fails to model the quadratic nature effectively, as evidenced by similar correlation metrics and negative R-squared. The Neighbor Similarity Analysis reveals that the best expression identified by the algorithm ($X^2$ + 10X) achieves a similarity of 0.4000 and fitness of 0.9197, outperforming its neighbors significantly. The average neighbor fitness is 0.0128, highlighting that the best expression represents a local optimum within the search space. This indicates that while the algorithm has progressed, further refinement or increased search depth may be required to capture the true underlying function fully.

**Advanced Explanation for $X^2$ + 10X:**
Expression Breakdown:
Term: X
  Fitness: -0.1698
  MSE Contribution: 11070.5892
  Complexity: 0
  Correlation: 0.9752
  Spearman Correlation: 1.0000
  Cosine Similarity: 0.0347
  Euclidean Similarity: -0.5301
  R-squared: -1.3412
  Relative Improvement: 0.0000

Term: X + 10
  Fitness: -0.0812
  MSE Contribution: 9530.5219
  Complexity: 1
  Correlation: 0.9752
  Spearman Correlation: 1.0000
  Cosine Similarity: 0.0395
  Euclidean Similarity: -0.4197
  R-squared: -1.0155
  Relative Improvement: 0.0000

**Neighbor Similarity Analysis:**
Neighbor 1:
  Expression: X*(X + 9)

```
 Similarity: 0.4000
 Fitness: 0.9197
Neighbor 2:
 Expression: X + 2
 Similarity: 0.0000
 Fitness: -0.1507
Neighbor 3:
 Expression: X + atan(X)
 Similarity: 0.0000
 Fitness: -0.1564
Neighbor 4:
 Expression: cos(X)
 Similarity: 0.0000
 Fitness: -0.3935
Neighbor 5:
 Expression: X + sqrt(2)
 Similarity: 0.0000
 Fitness: -0.1553

Insights:
Average Neighbor Fitness: 0.0128
Best Expression Fitness: 0.9436
The best expression outperforms its neighbors, suggesting it's a local optimum.
```

## 5.0 Limitations and future research needs

Despite its advantages, symbolic regression can suffer from significant challenges. For a start, the computational intensity required to evolve mathematical expressions, especially for large and high-dimensional datasets, can be a bottleneck [45]. Although advancements in parallel computing and algorithmic efficiency have alleviated some of these issues, optimizing symbolic regression for scalability and real-time applications continues to be a critical area of exploration. Additionally, the tendency of symbolic regression algorithms to generate overly complex models requires the development of regularization techniques and selection criteria that facilitate generalizability and prevent overfitting [46].

Another challenge is related to integrating domain knowledge into symbolic regression. Incorporating such knowledge, e.g., invariance properties, physical laws, or known functional forms, can constrain the search space and guide the evolutionary process [47]. When successful, these can enhance the relevance and interpretability of the generated expressions. On a similar front, integrating symbolic regression with data preprocessing and feature engineering can also enhance the extraction of meaningful patterns from complex datasets. Thus, research into embedding domain-specific constraints and synergizing symbolic regression with knowledge-driven approaches can be a promising avenue for developing more informed and context-aware algorithms. From the lens of the proposed algorithm, the authors would like to point out that there is a need to improve its performance in terms of calculation time with regard to the similarity metrics. Enabling advanced parallel computation or memonization could help speed the calculation processing of SPINEX.

## 6.0 Conclusions

In this study, we introduced SPINEX_SymbolicRegression, which extends the SPINEX (Similarity-based Predictions with Explainable Neighbors Exploration) framework. This algorithm employs a similarity-driven strategy to effectively identify high-fitness symbolic expressions by tracing both accuracy and structural similarity metrics. Extensive benchmarking was conducted, comparing SPINEX_SymbolicRegression against 182 mathematical functions sourced from international problem sets. These functions encompassed both randomly generated expressions and those derived from real-world physical phenomena. The performance evaluations focused on several critical aspects, including accuracy, expression similarity in terms of operators and variable inclusion, population dynamics, and convergence rates. The results demonstrate that SPINEX_SymbolicRegression consistently delivers competitive performance, frequently outperforming established leading algorithms.

## Data Availability

Some or all data, models, or code that support the findings of this study are available from the corresponding author upon reasonable request.

SPINEX can be accessed from [**to be added**].

## Conflict of Interest

The authors declare no conflict of interest.

# Appendix

```python
import numpy as np
import pandas as pd
from sklearn.metrics import mean_squared_error, r2_score
import matplotlib.pyplot as plt
from joblib import Parallel, delayed, parallel_backend
from sympy.printing.numpy import NumPyPrinter
from sympy.utilities.lambdify import lambdify
import sympy as sp
import warnings
from scipy.stats import pearsonr, spearmanr
from sklearn.preprocessing import StandardScaler
from scipy.spatial.distance import cosine, euclidean
import random
import threading
import time
import logging
import concurrent.futures
from functools import lru_cache
from sklearn.feature_selection import mutual_info_regression
import threading

class TimeoutMixin:
    def check_timeout(self):
        if hasattr(self, 'start_time') and hasattr(self, 'max_time'):
            if self.max_time is not None:
                elapsed = time.time() - self.start_time
                if elapsed > self.max_time:
                    raise TimeoutException(f"Maximum time of {self.max_time} seconds exceeded")

class TimeoutException(Exception):
    pass

def timeout_handler(timeout_duration, function, *args, **kwargs):
    result = [TimeoutException("Expression evaluation timed out")]
    def target():
        try:
            result[0] = function(*args, **kwargs)
        except Exception as e:
            result[0] = e
    thread = threading.Thread(target=target)
    thread.start()
    thread.join(timeout_duration)
    if thread.is_alive():
        thread.join()
        raise TimeoutException("Expression evaluation timed out")
    if isinstance(result[0], Exception):
        raise result[0]
    return result[0]

def set_seed(seed_value=42):
    np.random.seed(seed_value)
    random.seed(seed_value)

set_seed(42)

@lru_cache(maxsize=5000)
def cached_get_expression_depth_helper(expr):
    if isinstance(expr, (sp.Symbol, sp.Number)):
        return 1
    elif isinstance(expr, sp.Basic):
        if not expr.args:
            return 1
        return 1 + max(cached_get_expression_depth_helper(arg) for arg in expr.args)
    else:
```

```python
        return 1

def clear_all_lru_caches():
    cached_get_expression_depth_helper.cache_clear()

class SPINEX_SymbolicRegression(TimeoutMixin):
    def __init__(self, data, target, actual_function, user_expression=None, max_depth=None, population_size=50,
             generations=50, similarity_threshold=0.7, dynamic_similarity_threshold=True, n_jobs=1,
             explainability_level='none', plot_results=True, early_stopping_metric=None, early_stopping_value=None,
             patience=0, force_all_variables=False, variable_inclusion_strategy='guided', logger=None,
             last_resort=False, elite_percentage=0.05, dynamic_elite=True, max_time=None, relevance_metric='both'):
        self.max_time = max_time
        self.start_time = time.time()
        self._expression_depth_cache = {}
        self._evaluate_expression_cache = {}
        self._subexpressions_cache = {}
        self._tree_edit_distance_cache = {}
        self.tree_edit_distance_cache_lock = threading.Lock()
        self.similarity_cache = {}
        self.similarity_cache_lock = threading.Lock()
        self.lambdified_cache = {}
        self.data = data
        self.target = target
        self.actual_function = actual_function
        self.force_all_variables = force_all_variables
        self.variable_inclusion_strategy = variable_inclusion_strategy
        self.variables = {str(var): sp.Symbol(var) for var in list(data.columns)}
        if logger is None:
            self.logger = Logger()
        else:
            self.logger = logger
        base_depth = 2
        if max_depth is None:
            self.max_depth = base_depth + (len(self.variables) if self.force_all_variables else 2)
            self.logger.debug(f"max_depth set to {self.max_depth} based on {'all variables' if self.force_all_variables else 'base depth + 2'}")
        else:
            min_depth = base_depth + (len(self.variables) if self.force_all_variables else 2)
            if max_depth < min_depth:
                self.logger.warning(f"Provided max_depth={max_depth} is too low. Setting to minimum required: {min_depth}.")
                self.max_depth = min_depth
            else:
                self.max_depth = max_depth
                self.logger.debug(f"max_depth provided: {self.max_depth}")
        self.initial_max_depth = self.max_depth
        self.logger.debug(f"initial_max_depth set to {self.initial_max_depth}")
        self.relevance_metric = relevance_metric
        self.relevance_scores = self.assess_variable_relevance(relevance_metric=self.relevance_metric)
        self.sorted_vars = sorted(
            self.variables.values(),
            key=lambda var: self.relevance_scores.get(var, 0),
            reverse=True
        )
        self.logger.debug(f"Variables sorted by relevance: {self.sorted_vars}")
        self.user_expression = user_expression
        if self.user_expression:
            try:
                self.user_expression_sym = sp.sympify(self.user_expression)
                if self.get_expression_depth(self.user_expression_sym) > self.max_depth:
                    self.logger.warning("User expression exceeds max_depth. It will be truncated or modified.")
                    self.user_expression_sym = self.truncate_expression(self.user_expression_sym)
                allowed_symbols = set(self.variables.values())
                if not self.user_expression_sym.free_symbols.issubset(allowed_symbols):
                    self.logger.warning("User expression contains variables not present in the data. They will be removed.")
                    self.user_expression_sym = self.user_expression_sym.subs(
                        {var: 0 for var in self.user_expression_sym.free_symbols if var not in allowed_symbols}
                    )
```

```python
            except Exception as e:
                self.logger.error(f"Invalid user_expression provided: {e}")
                self.user_expression_sym = None
        else:
            self.user_expression_sym = None
        self.population_size = population_size
        self.generations = generations
        self.similarity_threshold = similarity_threshold
        self.initial_similarity_threshold = similarity_threshold
        self.dynamic_similarity_threshold = dynamic_similarity_threshold
        self.elite_percentage = elite_percentage
        self.dynamic_elite = dynamic_elite
        if not (0 < self.elite_percentage <= 1):
            raise ValueError("elite_percentage must be a float between 0 and 1")
        self.n_jobs = n_jobs
        self.explainability_level = explainability_level
        self.plot_results = plot_results
        self.early_stopping_metric = early_stopping_metric
        valid_metrics = ['fitness', 'r2', 'mse']
        if self.early_stopping_metric is not None:
            self.early_stopping_metric = self.early_stopping_metric.lower()
            if self.early_stopping_metric not in valid_metrics:
                raise ValueError(f"Invalid early_stopping_metric: {early_stopping_metric}. Choose from {valid_metrics}.")
            if early_stopping_value is None:
                raise ValueError("early_stopping_value must be provided when early_stopping_metric is set.")
        self.early_stopping_value = early_stopping_value
        self.patience = patience
        self.fitness_tolerance = 1e-6
        self.epsilon = 1e-10
        self.lambdified_cache = {}
        self.population = []
        self.best_expression = None
        self.baseline_mse = mean_squared_error(self.target, np.mean(self.target) * np.ones_like(self.target))
        self.expression_cache = {}
        self.max_cache_size = 1000  # Adjust based on available memory
        self.diversity_weight = 0.3  # Weight for balancing similarity and fitness
        self.mutation_types = ['replace', 'add', 'remove']
        valid_explain_levels = ['none', 'basic', 'advanced']
        if self.explainability_level.lower() not in valid_explain_levels:
            raise ValueError(f"Invalid explainability_level: {self.explainability_level}. "
                             f"Choose from {valid_explain_levels}.")
        self.unary_operators = [
            sp.sin, sp.cos,
            self.safe_exp, self.safe_log,
            self.safe_sqrt, self.safe_atan,
            self.safe_acos, self.safe_asin,
            self.safe_sinh, self.safe_cosh,
            self.safe_tanh
        ]
        self.binary_operators = [
            sp.Add,
            lambda x, y: x - y,
            sp.Mul,
            self.safe_div,
            self.safe_max,
            self.safe_min,
            self.safe_pow
        ]
        self.constant_choices = [sp.Integer(i) for i in range(-10, 11)] + \
                    [sp.pi, sp.E, sp.Rational(1, 2), sp.Rational(1, 3),
                     sp.Rational(1, 4), sp.Rational(1, 5), sp.sqrt(2), sp.sqrt(3)]
        self.variable_values = list(self.variables.values())
        self.variable_and_constant_choices = self.variable_values + self.constant_choices
        self.all_variables_set = set(self.variables.values())
        self.lambdify_modules = ['numpy', {
            'Max': np.maximum,
```

```python
            'Min': np.minimum,
            'sin': np.sin,
            'cos': np.cos,
            'tan': np.tan,
            'asin': np.arcsin,
            'acos': np.arccos,
            'atan': np.arctan,
            'sinh': np.sinh,
            'cosh': np.cosh,
            'tanh': np.tanh,
            'exp': np.exp,
            'sqrt': np.sqrt,
            'Abs': np.abs,
            'ceiling': np.ceil,
            'floor': np.floor
        }]
    self.unit_array = np.array([1.0])
    self.num_variables = len(self.variables)
    self.sample_value_arrays = [self.unit_array] * self.num_variables
    self.last_resort = last_resort
    self.last_resort_activated = False
    self.logger.info("SPINEX_SymbolicRegression initialized with parameters:")
    self.logger.info(f"max_depth={max_depth}, population_size={population_size}, generations={generations}, "
                     f"similarity_threshold={similarity_threshold}, n_jobs={n_jobs}, "
                     f"explainability_level={explainability_level}, plot_results={plot_results}, "
                     f"early_stopping_fitness={early_stopping_value}, patience={patience}, "
                     f"force_all_variables={force_all_variables}, variable_inclusion_strategy={variable_inclusion_strategy}")

def time_exceeded(self):
    if self.max_time is not None:
        elapsed_time = time.time() - self.start_time
        return elapsed_time > self.max_time
    else:
        return False

def get_expression_depth_cached(self, expr):
    if expr in self._expression_depth_cache:
        return self._expression_depth_cache[expr]
    depth = cached_get_expression_depth_helper(expr)
    self._expression_depth_cache[expr] = depth
    return depth

def get_expression_depth(self, expr):
    return self.get_expression_depth_cached(expr)

def clip_sympy(self, x, lower, upper):
    return sp.Piecewise(
        (lower, x < lower),
        (upper, x > upper),
        (x, True)
    )

def safe_log(self, x):
    is_sympy = isinstance(x, sp.Basic)
    if is_sympy:
        return sp.log(sp.Abs(x) + self.epsilon)
    else:
        x_array = np.asarray(x)
        return np.log(np.abs(x_array) + self.epsilon)

def safe_sqrt(self, x):
    is_sympy = isinstance(x, sp.Basic)
    if is_sympy:
        return sp.sqrt(sp.Abs(x))
    else:
        x_array = np.asarray(x)
```

```python
            return np.sqrt(np.abs(x_array))

    def safe_div(self, x, y):
        is_sympy = isinstance(x, sp.Basic) or isinstance(y, sp.Basic)
        small_value = 1e-12
        if is_sympy:
            return x / (y + small_value)
        else:
            y_array = np.asarray(y)
            y_array = np.where(np.abs(y_array) < small_value, small_value, y_array)
            return np.divide(x, y_array)

    def safe_max(self, x, y):
        is_sympy = isinstance(x, sp.Basic) or isinstance(y, sp.Basic)
        if is_sympy:
            return sp.Max(x, y)
        else:
            x_array = np.asarray(x)
            y_array = np.asarray(y)
            return np.maximum(x_array, y_array)

    def safe_min(self, x, y):
        is_sympy = isinstance(x, sp.Basic) or isinstance(y, sp.Basic)
        if is_sympy:
            return sp.Min(x, y)
        else:
            x_array = np.asarray(x)
            y_array = np.asarray(y)
            return np.minimum(x_array, y_array)

    def safe_pow(self, x, y):
        is_sympy = isinstance(x, sp.Basic) or isinstance(y, sp.Basic)
        if is_sympy:
            return sp.Pow(sp.Abs(x), y)
        else:
            x_array = np.abs(np.asarray(x))
            return np.power(x_array, y)

    def safe_asin(self, x):
        is_sympy = isinstance(x, sp.Basic)
        if is_sympy:
            return sp.asin(self.clip_sympy(x, -1, 1))
        else:
            x_array = np.asarray(x)
            return np.arcsin(np.clip(x_array, -1, 1))

    def safe_acos(self, x):
        is_sympy = isinstance(x, sp.Basic)
        if is_sympy:
            return sp.acos(self.clip_sympy(x, -1, 1))
        else:
            x_array = np.asarray(x)
            return np.arccos(np.clip(x_array, -1, 1))

    def safe_atan(self, x):
        is_sympy = isinstance(x, sp.Basic)
        if is_sympy:
            return sp.atan(x)
        else:
            x_array = np.asarray(x)
            return np.arctan(x_array)

    def safe_sinh(self, x):
        is_sympy = isinstance(x, sp.Basic)
        if is_sympy:
            return sp.sinh(self.clip_sympy(x, -100, 100))
```

```python
        else:
            x_array = np.asarray(x)
            return np.sinh(np.clip(x_array, -100, 100))

def safe_cosh(self, x):
    is_sympy = isinstance(x, sp.Basic)
    if is_sympy:
        return sp.cosh(self.clip_sympy(x, -100, 100))
    else:
        x_array = np.asarray(x)
        return np.cosh(np.clip(x_array, -100, 100))

def safe_tanh(self, x):
    is_sympy = isinstance(x, sp.Basic)
    if is_sympy:
        return sp.tanh(x)
    else:
        x_array = np.asarray(x)
        return np.tanh(x_array)

def safe_exp(self, x):
    is_sympy = isinstance(x, sp.Basic)
    if is_sympy:
        abs_x = sp.Abs(x)
        return sp.exp(abs_x / (abs_x + 1))
    else:
        x_array = np.abs(np.asarray(x))
        return np.exp(x_array / (x_array + 1))

def assess_variable_relevance(self, relevance_metric='correlation'):
    relevance_scores = {}
    if relevance_metric not in ['correlation', 'mutual_information', 'both']:
        raise ValueError("relevance_metric must be 'correlation', 'mutual_information', or 'both'.")
    if relevance_metric in ['mutual_information', 'both']:
        try:
            mi_scores = mutual_info_regression(
                self.data.values,
                self.target,
                discrete_features=False,
                random_state=42  # For reproducibility
            )
            mi_dict = dict(zip(self.variables.keys(), mi_scores))
        except Exception as e:
            self.logger.error(f"Error computing mutual information: {e}")
            mi_dict = {var: 0 for var in self.variables}
    else:
        mi_dict = {var: 0 for var in self.variables}
    for var in self.variables:
        try:
            var_values = self.data[var].values
            if relevance_metric in ['correlation', 'both']:
                pearson_corr, _ = pearsonr(var_values, self.target)
                pearson_corr = abs(pearson_corr)
            else:
                pearson_corr = 0
            if relevance_metric in ['mutual_information', 'both']:
                mi = mi_dict[var]
            else:
                mi = 0
            if relevance_metric == 'both':
                relevance_score = 0.5 * pearson_corr + 0.5 * mi
            elif relevance_metric == 'correlation':
                relevance_score = pearson_corr
            elif relevance_metric == 'mutual_information':
                relevance_score = mi
            relevance_scores[self.variables[var]] = relevance_score
```

```python
        except Exception as e:
            self.logger.error(f"Error assessing relevance for variable {var}: {e}")
            relevance_scores[self.variables[var]] = 0
    return relevance_scores

def initialize_population(self):
    self.check_timeout()
    if self.time_exceeded():
        self.logger.warning("Maximum time exceeded during population initialization. Returning with current population.")
        return
    self.logger.info("Initializing population...")
    self.population = []
    num_generated = self.population_size - int(self.user_expression_sym is not None)
    expr_types = np.random.choice(['simple', 'unary', 'binary', 'complex'], size=num_generated)
    for expr_type in expr_types:
        if self.time_exceeded():
            self.logger.warning("Maximum time exceeded during population initialization loop. Stopping population initialization.")
            break
        if expr_type == 'simple':
            expr = self.generate_leaf()
        elif expr_type == 'unary':
            expr = np.random.choice(self.unary_operators)(self.generate_leaf())
        elif expr_type == 'binary':
            expr = np.random.choice(self.binary_operators)(self.generate_leaf(), self.generate_leaf())
        else:
            expr = self.generate_random_expression()
        if expr is None:
            self.logger.warning("Generated expression is None. Skipping this expression.")
            continue
        if self.force_all_variables:
            if self.variable_inclusion_strategy == 'guided':
                missing_vars = set(self.variables.values()) - expr.free_symbols
                for var in self.sorted_vars:
                    if var in missing_vars:
                        potential_expr = sp.Add(expr, var)
                        if self.get_expression_depth(potential_expr) <= self.max_depth:
                            expr = potential_expr
                            missing_vars.remove(var)
            elif self.variable_inclusion_strategy == 'probabilistic':
                missing_vars = set(self.variables.values()) - expr.free_symbols
                for var in missing_vars:
                    if np.random.random() < 0.5:
                        potential_expr = sp.Add(expr, var)
                        if self.get_expression_depth(potential_expr) <= self.max_depth:
                            expr = potential_expr
        self.population.append(expr)
    if self.user_expression_sym:
        self.population.append(self.user_expression_sym)
        self.logger.info(f"User-provided expression added to population: {self.user_expression_sym}")
    self.logger.success(f"Population initialized with {len(self.population)} expressions.")
    if self.force_all_variables:
        self.logger.info(f"Variable inclusion strategy: {self.variable_inclusion_strategy}")

def generate_random_expression(self, depth=0, used_vars=None):
    self.check_timeout()
    if self.time_exceeded():
        self.logger.warning("Maximum time exceeded during random expression generation. Returning fallback expression.")
        return sp.Integer(1)
    if used_vars is None:
        used_vars = set()
    if depth == self.max_depth or (depth > 0 and np.random.random() < 0.3):
        if np.random.random() < 0.9:
            available_vars = list(set(self.variables.values()) - used_vars)
            if available_vars:
                expr = np.random.choice(available_vars)
            else:
```

```python
            expr = np.random.choice(self.variable_values)
            used_vars.add(expr)
        else:
            expr = np.random.choice(self.constant_choices)
        if self.force_all_variables:
            missing_vars = self.all_variables_set - used_vars
            for extra_var in missing_vars:
                potential_expr = sp.Add(expr, extra_var)
                if self.get_expression_depth(potential_expr) <= self.max_depth:
                    expr = potential_expr
                    used_vars.add(extra_var)
                else:
                    self.logger.debug(f"Cannot add variable {extra_var} to expression {expr} without exceeding max_depth.")
        return expr
    op_type = np.random.choice(['unary', 'binary'], p=[0.5, 0.5])
    max_attempts = 5
    for attempt in range(max_attempts):
        if self.time_exceeded():
            self.logger.warning("Maximum time exceeded during random expression generation loop. Returning fallback expression.")
            return sp.Integer(1)
        try:
            if op_type == 'unary':
                op = np.random.choice(self.unary_operators)
                new_subexpr = self.generate_random_expression(depth + 1, used_vars)
                if new_subexpr is None:
                    self.logger.debug("Generated sub-expression is None. Skipping this attempt.")
                    continue
                if depth + self.get_expression_depth(new_subexpr) > self.max_depth:
                    continue
                expr = op(new_subexpr)
            else:
                op = np.random.choice(self.binary_operators)
                left_expr = self.generate_random_expression(depth + 1, used_vars)
                right_expr = self.generate_random_expression(depth + 1, used_vars)
                if left_expr is None or right_expr is None:
                    self.logger.debug("One of the generated sub-expressions is None. Skipping this attempt.")
                    continue
                combined_depth = max(self.get_expression_depth(left_expr), self.get_expression_depth(right_expr)) + 1
                if combined_depth > self.max_depth:
                    continue
                expr = op(left_expr, right_expr)
                used_vars.update(expr.free_symbols)
            if self.is_valid_expression(expr):
                used_vars.update(expr.free_symbols)
                if self.user_expression_sym and np.random.random() < 0.1:
                    expr = self.insert_user_expression_subtree(expr)
                return expr
        except Exception as e:
            self.logger.error(f"Exception during expression generation attempt {attempt + 1}: {e}")
            continue
    fallback_expr = np.random.choice(self.variable_and_constant_choices)
    if self.force_all_variables:
        missing_vars = set(self.variables.values()) - {fallback_expr}
        for extra_var in missing_vars:
            potential_expr = sp.Add(fallback_expr, extra_var)
            if self.get_expression_depth(potential_expr) <= self.max_depth:
                fallback_expr = potential_expr
            else:
                self.logger.debug(f"Cannot add variable {extra_var} to fallback expression {fallback_expr} without exceeding max_depth.")
    return fallback_expr

def generate_leaf(self):
    return np.random.choice(list(self.variables.values()) +
                [sp.Integer(i) for i in range(-10, 11)] +
                [sp.pi, sp.E, sp.Rational(1, 2), sp.Rational(1, 3),
                 sp.Rational(1, 4), sp.Rational(1, 5), sp.sqrt(2), sp.sqrt(3)])
```

```python
def is_valid_expression(self, expr):
    self.check_timeout()
    try:
        if expr in self.expression_cache:
            return True
        func = self.get_lambdified_function(expr)
        if func is None:
            return False
        predictions = func(*self.sample_value_arrays)
        result = predictions[0]
        is_valid = (not np.iscomplex(result) and
                    not np.isnan(result) and
                    not np.isinf(result))
        return is_valid
    except Exception as e:
        return False

def dynamic_complexity_penalty(self, complexity, generation, best_fitness):
    if self.last_resort_activated:
        final_penalty = 1.0
    else:
        progress = generation / self.generations
        base_penalty = 0.01 * (1 + np.tanh(2 * progress - 1))
        fitness_factor = np.clip(1.5 - (best_fitness * 1.5), 0.5, 1.5)
        complexity_scale = np.log1p(complexity) / np.log1p(self.max_depth)
        unique_exprs = len(set(str(expr) for expr in self.population))
        diversity = unique_exprs / len(self.population)
        diversity_factor = 1 + np.tanh(5 * (diversity - 0.5))
        final_penalty = np.exp(-base_penalty * fitness_factor * complexity_scale * diversity_factor)
        final_penalty = np.clip(final_penalty, 0.1, 1.0)
    return final_penalty

def evaluate_expression(self, expression, timeout=5):
    self.check_timeout()
    try:
        if isinstance(expression, str):
            expression = sp.sympify(expression)
        if (expression.has(sp.zoo) or
            expression.has(sp.oo) or
            expression.has(sp.S.NegativeInfinity) or
            expression.has(sp.nan)):
            raise ValueError(f"Invalid expression contains infinity or NaN: {expression}")
        func = self.get_lambdified_function(expression)
        if func is None:
            raise ValueError(f"Could not lambdify expression: {expression}")
        def evaluation_function():
            return func(*[self.data[str(var)].values for var in self.variables])
        predictions = timeout_handler(timeout, evaluation_function)
        if np.isscalar(predictions):
            predictions = np.full_like(self.target, predictions)
        if np.isnan(predictions).any() or np.isinf(predictions).any():
            return 0, np.inf, np.inf, -1, -1, -1, -1, -1, -1
        mse = mean_squared_error(self.target, predictions)
        r2 = r2_score(self.target, predictions)
        correlation, _ = pearsonr(self.target, predictions)
        spearman, _ = spearmanr(self.target, predictions)
        scaler = StandardScaler()
        target_normalized = scaler.fit_transform(self.target.reshape(-1, 1)).flatten()
        predictions_normalized = scaler.transform(predictions.reshape(-1, 1)).flatten()
        cosine_sim = 1 - cosine(target_normalized, predictions_normalized)
        euclidean_dist = euclidean(target_normalized, predictions_normalized)
        max_euclidean = np.sqrt(len(self.target))
        euclidean_sim = 1 - (euclidean_dist / max_euclidean)
        complexity = expression.count_ops()
        relative_improvement = max(0, (self.baseline_mse - mse) / self.baseline_mse)
```

```python
            accuracy_score = (0.2 * r2 +
                       0.2 * relative_improvement +
                       0.2 * cosine_sim +
                       0.2 * euclidean_sim +
                       0.1 * abs(correlation) +
                       0.1 * abs(spearman))
            complexity_penalty = self.dynamic_complexity_penalty(complexity, self.current_generation, self.best_fitness_overall)
            fitness = accuracy_score * complexity_penalty
            return fitness, mse, complexity, correlation, spearman, cosine_sim, euclidean_sim, r2, relative_improvement
        except TimeoutException:
            self.logger.error(f"Evaluation of expression {expression} timed out after {timeout} seconds")
            return 0, np.inf, np.inf, -1, -1, -1, -1, -1, -1
        except Exception as e:
            self.logger.error(f"Error evaluating expression: {expression} - {str(e)}")
            return 0, np.inf, np.inf, -1, -1, -1, -1, -1, -1

    def insert_user_expression_subtree(self, expr):
        if not self.user_expression_sym:
            return expr
        try:
            subexpr = self.get_random_subexpression(expr)
            new_expr = expr.subs(subexpr, self.user_expression_sym)
            if self.get_expression_depth(new_expr) <= self.max_depth:
                return new_expr
            else:
                self.logger.debug(f"Replacing with user expression exceeds max_depth. Skipping.")
                return expr
        except Exception as e:
            self.logger.debug(f"Error inserting user expression subtree: {e}")
            return expr

    def truncate_expression(self, expr):
        if self.get_expression_depth(expr) <= self.max_depth:
            return expr
        if isinstance(expr, sp.Basic):
            truncated_args = []
            for arg in expr.args:
                truncated_arg = self.truncate_expression(arg)
                truncated_args.append(truncated_arg)
                if self.get_expression_depth(sp.Function(expr.func, *truncated_args)) > self.max_depth:
                    break
            return expr.func(*truncated_args)
        return expr

    def cached_evaluate_expression(self, expression_str, timeout=5):
        if expression_str in self._evaluate_expression_cache:
            return self._evaluate_expression_cache[expression_str]
        expression = sp.sympify(expression_str)
        result = self.evaluate_expression(expression, timeout)
        self._evaluate_expression_cache[expression_str] = result
        return result

    def prune_population(self):
        self.check_timeout()
        self.population = [expr for expr in self.population if self.is_valid_expression(expr)]
        unique_exprs = {}
        for expr in self.population:
            expr_str = str(expr)
            if expr_str not in unique_exprs:
                unique_exprs[expr_str] = expr
        self.population = list(unique_exprs.values())
        if self.force_all_variables:
            self.population = [
                expr for expr in self.population
                if set(self.variables.values()).issubset(expr.free_symbols)
            ]
```

```python
            if len(self.population) > self.population_size:
                with parallel_backend('threading', n_jobs=self.n_jobs):
                    fitness_scores = Parallel()(delayed(self.cached_evaluate_expression)(str(expr), timeout=5) for expr in self.population)
                combined = list(zip(self.population, fitness_scores))
                sorted_population = sorted(combined, key=lambda x: x[1][0], reverse=True)
                self.population = [expr for expr, _ in sorted_population[:self.population_size]]
            if len(self.expression_cache) > self.max_cache_size:
                sorted_cache = sorted(self.expression_cache.items(), key=lambda x: x[1][0], reverse=True)
                self.expression_cache = dict(sorted_cache[:self.max_cache_size])

    def select_diverse_expression(self, population):
        if len(population) <= 1:
            return population[0][0]
        similarities = Parallel(n_jobs=self.n_jobs)(
            delayed(self.calculate_similarity)(expr1, expr2)
            for expr1, _ in population
            for expr2, _ in population
            if expr1 != expr2
        )
        avg_similarities = np.array(similarities).reshape(len(population), -1).mean(axis=1)
        diversity_scores = 1 - avg_similarities
        fitness_scores = np.array([score[0] for _, score in population])
        combined_scores = (1 - self.diversity_weight) * fitness_scores + self.diversity_weight * diversity_scores
        combined_scores /= combined_scores.sum()
        selected_index = np.random.choice(len(population), p=combined_scores)
        return population[selected_index][0]

    def select_similar_expression(self, target_expr, population):
        similarities = [
            (expr, self.calculate_similarity(target_expr, expr))
            for expr, _ in population if expr != target_expr
        ]
        similar_exprs = [expr for expr, sim in similarities if sim >= self.similarity_threshold]
        if similar_exprs:
            return np.random.choice(similar_exprs)
        else:
            return np.random.choice([expr for expr, _ in population if expr != target_expr])

    def mutate_expression(self, expression):
        self.check_timeout()
        if self.time_exceeded():
            raise TimeoutException("Maximum time exceeded during mutation.")
        if not isinstance(expression, sp.Basic):
            expr = sp.sympify(expression)
        else:
            expr = expression
        expr_depth = self.get_expression_depth(expr)
        mutation_type = np.random.choice(['replace', 'add', 'remove'])
        mutated_expr = None
        max_attempts = 5
        for attempt in range(max_attempts):
            if self.time_exceeded():
                raise TimeoutException("Maximum time exceeded during mutation.")
            try:
                if mutation_type == 'replace':
                    subexpr = self.get_random_subexpression(expr)
                    new_subexpr = self.generate_random_expression()
                    potential_depth = expr_depth - self.get_expression_depth(subexpr) + self.get_expression_depth(new_subexpr)
                    if potential_depth > self.max_depth:
                        continue
                    mutated_expr = expr.subs(subexpr, new_subexpr)
                elif mutation_type == 'add':
                    new_expr = self.generate_random_expression()
                    combined_depth = max(expr_depth, self.get_expression_depth(new_expr)) + 1
                    if combined_depth > self.max_depth:
                        continue
```

```
                op = np.random.choice(self.binary_operators)
                mutated_expr = op(expr, new_expr)
            elif mutation_type == 'remove':
                if len(expr.args) > 1 or isinstance(expr.func, (sp.Add, sp.Mul)):
                    args = list(expr.args)
                    args.pop(np.random.randint(len(args)))
                    if isinstance(expr.func, sp.Pow) and len(args) < 2:
                        continue
                    temp_expr = expr.func(*args)
                    temp_expr_depth = self.get_expression_depth(temp_expr)
                    if temp_expr_depth > self.max_depth:
                        continue
                    mutated_expr = temp_expr
                else:
                    continue
            else:
                continue
            if mutated_expr != expr:
                similarity = self.calculate_similarity(expr, mutated_expr)
                if similarity < 0.9:
                    if self.force_all_variables:
                        missing_vars = set(self.variables.values()) - mutated_expr.free_symbols
                        if self.variable_inclusion_strategy == 'guided':
                            for var in self.sorted_vars:
                                if var in missing_vars and np.random.random() < 0.5:
                                    potential_expr = sp.Add(mutated_expr, var)
                                    if self.get_expression_depth(potential_expr) <= self.max_depth:
                                        mutated_expr = potential_expr
                                        missing_vars.remove(var)
                        elif self.variable_inclusion_strategy == 'probabilistic':
                            for var in missing_vars:
                                if np.random.random() < 0.5:
                                    potential_expr = sp.Add(mutated_expr, var)
                                    if self.get_expression_depth(potential_expr) <= self.max_depth:
                                        mutated_expr = potential_expr
                    return mutated_expr
        except Exception as e:
            continue
    return self.generate_random_expression()

def crossover_expressions(self, expr1, expr2):
    self.check_timeout()
    if self.time_exceeded():
        raise TimeoutException("Maximum time exceeded during crossover.")
    tree1 = sp.sympify(expr1)
    tree2 = sp.sympify(expr2)
    max_attempts = 5
    new_expr = None
    for attempt in range(max_attempts):
        if self.time_exceeded():
            raise TimeoutException("Maximum time exceeded during crossover.")
        try:
            subtree1 = self.get_random_subexpression(tree1)
            subtree2 = self.get_random_subexpression(tree2)
            depth1 = self.get_expression_depth(subtree1)
            depth2 = self.get_expression_depth(subtree2)
            expr1_depth = self.get_expression_depth(tree1)
            expr2_depth = self.get_expression_depth(tree2)
            new_depth1 = expr1_depth - depth1 + depth2
            new_depth2 = expr2_depth - depth2 + depth1
            if new_depth1 > self.max_depth or new_depth2 > self.max_depth:
                continue
            new_tree1 = tree1.subs(subtree1, subtree2)
            new_tree2 = tree2.subs(subtree2, subtree1)
            new_expr = np.random.choice([new_tree1, new_tree2])
            if new_expr != tree1 and new_expr != tree2 and self.calculate_similarity(tree1, new_expr) < 0.9:
```

```python
            if self.force_all_variables:
                missing_vars = set(self.variables.values()) - new_expr.free_symbols
                if self.variable_inclusion_strategy == 'guided':
                    for var in self.sorted_vars:
                        if var in missing_vars and np.random.random() < 0.5:
                            potential_expr = sp.Add(new_expr, var)
                            if self.get_expression_depth(potential_expr) <= self.max_depth:
                                new_expr = potential_expr
                                missing_vars.remove(var)
                elif self.variable_inclusion_strategy == 'probabilistic':
                    for var in missing_vars:
                        if np.random.random() < 0.5:
                            potential_expr = sp.Add(new_expr, var)
                            if self.get_expression_depth(potential_expr) <= self.max_depth:
                                new_expr = potential_expr
            return new_expr
        except Exception as e:
            continue
    return self.generate_random_expression()

def get_all_subexpressions_cached(self, expr_str):
    if expr_str in self._subexpressions_cache:
        return self._subexpressions_cache[expr_str]
    expr = sp.sympify(expr_str)
    subexprs = self.get_all_subexpressions_helper(expr)
    self._subexpressions_cache[expr_str] = subexprs
    return subexprs

def get_all_subexpressions_helper(self, expr):
    if isinstance(expr, (sp.Symbol, sp.Number)):
        return [expr]
    subexprs = [expr]
    for arg in expr.args:
        subexprs.extend(self.get_all_subexpressions_helper(arg))
    return subexprs

def get_random_subexpression(self, expr):
    expr_str = str(expr)
    subexprs = self.get_all_subexpressions_cached(expr_str)
    return np.random.choice(subexprs)

def get_lambdified_function(self, expression):
    try:
        if expression in self.lambdified_cache:
            return self.lambdified_cache[expression]
        func = sp.lambdify(
            self.variable_values,
            expression,
            modules=self.lambdify_modules
        )
        self.lambdified_cache[expression] = func
        return func
    except Exception as e:
        self.logger.error(f"Error in get_lambdified_function: {e}")
        return None

def update_similarity_threshold(self):
    unique_expressions = set(str(expr) for expr in self.population)
    diversity = len(unique_expressions) / len(self.population)
    self.diversity = diversity
    if not self.dynamic_similarity_threshold:
        return
    self.similarity_threshold = max(0.1, min(0.9, self.initial_similarity_threshold * (1 - diversity)))
    self.logger.info(f"Updated similarity_threshold to {self.similarity_threshold:.2f} based on diversity {diversity:.2f}")

def update_elite_percentage(self):
```

```python
        if not self.dynamic_elite:
            return
        diversity = self.diversity
        progress = self.current_generation / self.generations
        base_elite = 0.05
        max_elite = 0.2
        self.elite_percentage = min(max_elite, base_elite + (progress * (1 - diversity) * (max_elite - base_elite)))
        self.elite_percentage = max(base_elite, self.elite_percentage)
        self.logger.info(f"Updated elite_percentage to {self.elite_percentage:.4f} based on diversity {diversity:.4f} and progress {progress:.4f}")

    def evolve(self):
        self.check_timeout()
        self.start_time = time.time()
        self._expression_depth_cache.clear()
        self._evaluate_expression_cache.clear()
        self._subexpressions_cache.clear()
        self._tree_edit_distance_cache.clear()
        self.similarity_cache.clear()
        self.lambdified_cache.clear()
        set_seed(42)
        self.logger.info("Starting evolution process...")
        self.best_fitness_overall = float('-inf')
        self.best_r2_overall = float('-inf')
        self.best_mse_overall = float('inf')
        best_expression_overall = None
        stagnant_generations = 0
        reinitialization_attempts = 0
        max_reinitializations = int(0.9 * self.generations)
        self.current_generation = -1
        MAX_ALLOWED_DEPTH = 2 * self.initial_max_depth
        self.logger.info(f"Set MAX_ALLOWED_DEPTH to {MAX_ALLOWED_DEPTH} (twice the initial max_depth).")
        self.early_stopping_counter = 0
        for generation in range(self.generations):
            elapsed_time = time.time() - self.start_time
            if self.max_time is not None and elapsed_time > self.max_time:
                self.logger.warning(f"Maximum time of {self.max_time} seconds exceeded at generation {generation}. Terminating evolution early.")
                if best_expression_overall is not None:
                    self.best_expression = best_expression_overall
                    self.logger.info(f"Returning best expression found so far: {self.best_expression}")
                else:
                    self.logger.warning(f"Evolution completed at generation {generation} without an expression. Returning a constant fallback expression. Please consider updating your search settings.")
                    self.best_expression = sp.Integer(1)
                self.logger.success(
                    f"Evolution terminated due to timeout at generation {generation}. "
                    f"Best Expression: {self.best_expression}, "
                    f"R²: {self.best_r2_overall:.3g}, "
                    f"MSE: {self.best_mse_overall:.3g}, "
                    f"Fitness: {self.best_fitness_overall:.3g}"
                )
                return self.best_expression
            self.current_generation = generation
            max_depth_exceeded = False
            if (self.last_resort and not self.last_resort_activated and
                generation >= int(0.9 * self.generations) and
                self.best_fitness_overall < 0.85):
                self.activate_last_resort_mode()
            self.logger.info(f"--- Generation {generation} ---")
            try:
                with parallel_backend('threading', n_jobs=self.n_jobs):
                    fitness_scores = Parallel()(
                        delayed(self.cached_evaluate_expression)(str(expr), timeout=5) for expr in self.population
                    )
            except Exception as e:
                self.logger.error(f"Error during parallel fitness evaluation at generation {generation}: {e}")
                continue
```

```python
            valid_population = [
                (expr, score) for expr, score in zip(self.population, fitness_scores) if score[0] > 1e-10
            ]
            if not valid_population:
                self.logger.warning(f"Generation {generation}: No valid expressions found. Reinitializing population.")
                reinitialization_attempts += 1
                if reinitialization_attempts > max_reinitializations:
                    self.logger.error(f"Max reinitializations reached. Terminating evolution after {reinitialization_attempts} attempts.")
                    break
                self.initialize_population()
                continue
            reinitialization_attempts = 0
            if self.force_all_variables:
                sorted_population = sorted(
                    valid_population,
                    key=lambda x: (set(self.variables.values()).issubset(x[0].free_symbols), x[1][0]),
                    reverse=True
                )
            else:
                sorted_population = sorted(valid_population, key=lambda x: x[1][0], reverse=True)
            current_best_expression = sorted_population[0][0]
            current_best_fitness = sorted_population[0][1][0]
            current_best_mse = sorted_population[0][1][1]
            current_best_r2 = sorted_population[0][1][7]
            if current_best_fitness > self.best_fitness_overall:
                self.best_fitness_overall = current_best_fitness
                best_expression_overall = current_best_expression
                self.best_r2_overall = current_best_r2
                self.best_mse_overall = current_best_mse
                stagnant_generations = 0
                self.logger.info(f"New best fitness: {self.best_fitness_overall:.3g} with expression: {best_expression_overall}")
            else:
                stagnant_generations += 1
                self.logger.info(f"No improvement in fitness. Stagnant generations: {stagnant_generations}")
            self.logger.info(f"Generation {generation}: Best Fitness = {self.best_fitness_overall:.3g}: Best Expression: {best_expression_overall}")
            if self.early_stopping_metric is not None and self.early_stopping_value is not None:
                metric_name = self.early_stopping_metric.lower()
                metric_value = None
                comparison = False
                if metric_name == 'fitness':
                    metric_value = current_best_fitness
                    target_value = self.early_stopping_value
                    comparison = metric_value >= target_value - self.fitness_tolerance
                elif metric_name == 'r2':
                    metric_value = current_best_r2
                    target_value = self.early_stopping_value
                    comparison = metric_value >= target_value
                elif metric_name == 'mse':
                    metric_value = current_best_mse
                    target_value = self.early_stopping_value
                    comparison = metric_value <= target_value
                else:
                    raise ValueError(f"Invalid early_stopping_metric: {self.early_stopping_metric}")
                if comparison:
                    self.early_stopping_counter += 1
                    if self.early_stopping_counter >= self.patience:
                        self.logger.info(
                            f"Early stopping at generation {generation} as {metric_name} "
                            f"has reached the target value of {target_value} for {self.patience} consecutive generations."
                        )
                        break
                else:
                    self.early_stopping_counter = 0
            else:
                self.early_stopping_counter = 0
            if stagnant_generations > 5:
```

```python
                    self.logger.warning("Stagnation detected, resetting population.")
                    self.initialize_population()
                    stagnant_generations = 0
                    continue
            if self.last_resort_activated:
                mutation_rate = self.mutation_rate
            else:
                mutation_rate = 0.2 if stagnant_generations < 5 else 0.7
            self.logger.info(f"Mutation rate set to {mutation_rate} for generation {generation}.")
            self.update_similarity_threshold()
            self.update_elite_percentage()
            elite_count = max(1, int(self.elite_percentage * self.population_size))
            self.logger.info(f"Elitism: Carrying forward top {elite_count} expressions "
                             f"({self.elite_percentage * 100:.1f}% of population).")
            new_population = [item[0] for item in sorted_population[:elite_count]]
            max_generation_attempts = 10
            generation_attempts = 0
            while len(new_population) < self.population_size:
                generation_attempts += 1
                self.logger.debug(f"Generation attempts: {generation_attempts}")
                if generation_attempts > max_generation_attempts:
                    self.logger.warning(f"Max generation attempts reached. Generating random expressions to complete population.")
                    remaining_slots = self.population_size - len(new_population)
                    max_final_attempts = remaining_slots * 2
                    final_attempts = 0
                    exceeded_depth_count = 0
                    while len(new_population) < self.population_size and final_attempts < max_final_attempts:
                        expr = self.generate_random_expression()
                        if expr is None:
                            self.logger.warning("Generated expression is None during final attempts. Skipping.")
                            final_attempts += 1
                            continue
                        expr_depth = self.get_expression_depth(expr)
                        if expr_depth <= self.max_depth:
                            new_population.append(expr)
                            self.logger.debug(f"Added new expression within max_depth: {expr}")
                        else:
                            exceeded_depth_count += 1
                            self.logger.debug(f"Random expression {expr} exceeds max_depth ({expr_depth} > {self.max_depth}). Skipping.")
                        final_attempts += 1
                    exceed_threshold = 0.5
                    if final_attempts > 0 and (exceeded_depth_count / final_attempts) > exceed_threshold:
                        self.max_depth += 1
                        self.logger.warning(f"More than {exceed_threshold*100}% of expressions exceeded max_depth. Incrementing max_depth to {self.max_depth}.")
                        if self.max_depth > MAX_ALLOWED_DEPTH:
                            self.logger.error(f"Max depth exceeded the allowed limit of {MAX_ALLOWED_DEPTH}. Terminating evolution.")
                            max_depth_exceeded = True
                            break
                    if max_depth_exceeded:
                        break
                    if len(new_population) < self.population_size:
                        self.logger.warning(f"Only {len(new_population)} expressions were added after {final_attempts} attempts.")
                    break
                if np.random.random() < mutation_rate:
                    parent = np.random.choice([expr for expr, _ in sorted_population])
                    child = self.mutate_expression(parent)
                    if child is None:
                        self.logger.warning("Mutated child is None. Skipping.")
                        continue
                    self.logger.debug(f"Mutated expression: {child}")
                else:
                    parent = np.random.choice([expr for expr, _ in sorted_population])
                    similar_expressions = self.find_similar_expressions(parent)
                    if similar_expressions and np.random.random() > 0.5:
                        partner = np.random.choice(similar_expressions)
```

```
                    child = self.crossover_expressions(parent, partner)
                    if child is None:
                        self.logger.warning("Crossover resulted in None expression. Skipping.")
                        continue
                    self.logger.debug(f"Crossover between {parent} and {partner} resulted in {child}")
                else:
                    child = self.generate_random_expression()
                    if child is None:
                        self.logger.warning("Generated child is None during crossover/generation. Skipping.")
                        continue
                    self.logger.debug(f"Generated random expression: {child}")
                if self.force_all_variables:
                    missing_vars = set(self.variables.values()) - child.free_symbols
                    if self.variable_inclusion_strategy == 'guided':
                        for var in self.sorted_vars:
                            if var in missing_vars and np.random.random() < 0.5:
                                potential_expr = sp.Add(child, var)
                                if self.get_expression_depth(potential_expr) <= self.max_depth:
                                    child = potential_expr
                                    missing_vars.remove(var)
                    elif self.variable_inclusion_strategy == 'probabilistic':
                        for var in missing_vars:
                            if np.random.random() < 0.5:
                                potential_expr = sp.Add(child, var)
                                if self.get_expression_depth(potential_expr) <= self.max_depth:
                                    child = potential_expr
                if self.force_all_variables:
                    missing_vars = set(self.variables.values()) - child.free_symbols
                    for extra_var in missing_vars:
                        child = sp.Add(child, extra_var)
                if self.get_expression_depth(child) > self.max_depth:
                    self.logger.debug(f"Child expression {child} exceeds max_depth. Skipping.")
                    continue
                new_population.append(child)
            self.population = new_population
            self.logger.info(f"Population updated for generation {generation}.")
            if max_depth_exceeded:
                self.logger.error(f"Terminating evolution due to max_depth exceeding {MAX_ALLOWED_DEPTH}.")
                break
            self.prune_population()
            self.check_diversity()
        if best_expression_overall is None:
            self.logger.warning(f"Evolution completed at generation {self.current_generation} without an expression. "
                "Returning a constant fallback expression. Please consider updating your search settings.")
            best_expression_overall = sp.Integer(1)
        self.best_expression = best_expression_overall
        if self.force_all_variables:
            missing_vars = set(self.variables.values()) - self.best_expression.free_symbols
            if missing_vars:
                causes = self.determine_missing_variables_cause(missing_vars)
                self.logger.warning(
                    f"Best expression is missing variables: {missing_vars}. "
                    f"Possible cause(s): {causes}. "
                    "The algorithm will still return this expression, but you may want to adjust your settings."
                )
        elapsed_time = time.time() - self.start_time
        self.logger.success(
            f"Evolution completed at generation {self.current_generation}. "
            f"Time taken: {elapsed_time:.2f} seconds. "
            f"Best Expression: {self.best_expression}, "
            f"R²: {self.best_r2_overall:.3g}, "
            f"MSE: {self.best_mse_overall:.3g}, "
            f"Fitness: {self.best_fitness_overall:.3g}"
        )
        return self.best_expression
```

```python
def determine_missing_variables_cause(self, missing_vars):
    causes = []
    if self.get_expression_depth(self.best_expression) >= self.max_depth:
        causes.append(f"Expression reached max depth ({self.max_depth})")
    if self.variable_inclusion_strategy == 'probabilistic':
        causes.append("Probabilistic inclusion strategy may have excluded variables")
    if len(self.population) < self.population_size:
        causes.append("Population size decreased during evolution")
    if self.last_resort_activated:
        causes.append("Last resort mode was activated")
    if not causes:
        causes.append("Unknown - consider increasing generations or adjusting other parameters")
    return ", ".join(causes)

def check_diversity(self):
    unique_expressions = set(str(expr) for expr in self.population)
    diversity = len(unique_expressions) / len(self.population)
    if diversity < 0.5:
        self.logger.warning("Low diversity detected, introducing new random expressions.")
        num_new = int(self.population_size * 0.2)
        new_expressions = Parallel(n_jobs=self.n_jobs)(
            delayed(self.generate_random_expression)() for _ in range(num_new)
        )
        self.population = self.population[:-num_new] + new_expressions

def calculate_similarity(self, expr1, expr2):
    key = tuple(sorted([expr1, expr2], key=lambda x: x.sort_key()))
    similarity = self.tree_edit_distance(expr1, expr2)
    similarity = max(0.0, min(1.0, similarity))
    return similarity

def tree_edit_distance(self, tree1_str, tree2_str, depth=0):
    self.check_timeout()
    key = (tree1_str, tree2_str)
    with self.tree_edit_distance_cache_lock:
        if key in self._tree_edit_distance_cache:
            return self._tree_edit_distance_cache[key]
    tree1 = sp.sympify(tree1_str)
    tree2 = sp.sympify(tree2_str)
    if depth > self.max_depth:
        similarity = 1
    else:
        def _size(tree):
            if isinstance(tree, (sp.Symbol, sp.Number)):
                return 1
            return 1 + sum(_size(arg) for arg in tree.args)

        def _distance(t1, t2, depth):
            self.check_timeout()
            if t1 == t2:
                return 0
            if isinstance(t1, (sp.Symbol, sp.Number)) and isinstance(t2, (sp.Symbol, sp.Number)):
                return 1
            if isinstance(t1, (sp.Symbol, sp.Number)) or isinstance(t2, (sp.Symbol, sp.Number)):
                return max(_size(t1), _size(t2))
            if t1.func != t2.func:
                return 1 + max(_size(t1), _size(t2))
            return 1 + sum(_distance(a1, a2, depth + 1) for a1, a2 in zip(t1.args, t2.args))
        max_size = max(_size(tree1), _size(tree2))
        distance = _distance(tree1, tree2, depth)
        similarity = 1 - (distance / max_size)
        similarity = max(0.0, min(1.0, similarity))
    with self.tree_edit_distance_cache_lock:
        self._tree_edit_distance_cache[key] = similarity
    return similarity
```

```python
def find_similar_expressions(self, expression):
    with concurrent.futures.ThreadPoolExecutor() as executor:
        similarities = list(executor.map(lambda expr: self.calculate_similarity(expression, expr), self.population))
    return [expr for expr, sim in zip(self.population, similarities) if sim >= self.similarity_threshold]

def merge_expressions(self, expr1, expr2):
    tree1 = sp.sympify(expr1)
    tree2 = sp.sympify(expr2)
    operators = [sp.Add, sp.Mul]
    chosen_op = np.random.choice(operators)
    try:
        merged = chosen_op(tree1, tree2)
        merged_expr = sp.sympify(merged)
        if self.get_expression_depth(merged_expr) > self.max_depth:
            return self.generate_random_expression()
        if self.force_all_variables:
            missing_vars = set(self.variables.values()) - merged_expr.free_symbols
            for extra_var in missing_vars:
                merged_expr = sp.Add(merged_expr, extra_var)
        return merged_expr
    except Exception as e:
        self.logger.debug(f"Error merging expressions {expr1} and {expr2}: {e}")
        return self.generate_random_expression()

def activate_last_resort_mode(self):
    self.logger.warning("Activating Last Resort Mode: Increasing complexity and depth.")
    self.max_depth += 5
    self.mutation_rate = 0.85
    self.diversity_weight = 0.1
    self.last_resort_activated = True

def find_similar_neighbors(self, expression, n_neighbors=5):
    similarities = []
    for expr in self.population:
        if expr != expression:
            similarity = self.calculate_similarity(expression, expr)
            similarities.append((expr, similarity))
    return sorted(similarities, key=lambda x: x[1], reverse=True)[:n_neighbors]

def analyze_neighbors(self, expression, neighbors):
    analysis = []
    for neighbor, similarity in neighbors:
        neighbor_fitness, *_ = self.evaluate_expression(neighbor)
        analysis.append({
            'expression': str(neighbor),
            'similarity': similarity,
            'fitness': neighbor_fitness
        })
    return analysis

def explain(self):
    if self.explainability_level == 'basic':
        self.basic_explanation()
    elif self.explainability_level == 'advanced':
        self.advanced_explanation()
    elif self.explainability_level.lower() == 'none':
        if self.plot_results:
            self.plot_expression()
    else:
        self.logger.warning(f"Unknown explainability level: {self.explainability_level}. No explanation will be provided.")

def basic_explanation(self):
    print("\nBasic Explanation:")
    print(f"Best Expression: {self.best_expression}")
    fitness, mse, complexity, correlation, spearman, cosine_sim, euclidean_sim, r2, relative_improvement = self.evaluate_expression(self.best_expression)
```

```python
        print(f"Fitness: {fitness}")
        print(f"Mean Squared Error: {mse}")
        print(f"R-squared: {r2}")
        print(f"Relative Improvement: {relative_improvement}")
        print(f"Expression Complexity: {complexity}")
        print(f"Correlation: {correlation}")
        print(f"Spearman Correlation: {spearman}")
        print(f"Cosine Similarity: {cosine_sim}")
        print(f"Euclidean Similarity: {euclidean_sim}")
        self.plot_expression()

    def advanced_explanation(self):
        self.basic_explanation()
        print("\nAdvanced Explanation:")
        expr = sp.sympify(self.best_expression)
        print("Expression Breakdown:")
        for term in expr.args:
            term_expr = str(term)
            if isinstance(term, sp.Number):
                print(f"Term: {term_expr}")
                print(f"  Contribution: Constant term")
            else:
                term_fitness, term_mse, term_complexity, term_corr, term_spearman, term_cosine, term_euclidean, term_r2, term_rel_imp = self.evaluate_expression(term_expr)
                print(f"Term: {term_expr}")
                print(f"  Fitness: {term_fitness:.4f}")
                print(f"  MSE Contribution: {term_mse:.4f}")
                print(f"  Complexity: {term_complexity}")
                print(f"  Correlation: {term_corr:.4f}")
                print(f"  Spearman Correlation: {term_spearman:.4f}")
                print(f"  Cosine Similarity: {term_cosine:.4f}")
                print(f"  Euclidean Similarity: {term_euclidean:.4f}")
                print(f"  R-squared: {term_r2:.4f}")
                print(f"  Relative Improvement: {term_rel_imp:.4f}")
        print("\nNeighbor Similarity Analysis:")
        similar_neighbors = self.find_similar_neighbors(self.best_expression)
        neighbor_analysis = self.analyze_neighbors(self.best_expression, similar_neighbors)
        for i, neighbor in enumerate(neighbor_analysis, 1):
            print(f"Neighbor {i}:")
            print(f"  Expression: {neighbor['expression']}")
            print(f"  Similarity: {neighbor['similarity']:.4f}")
            print(f"  Fitness: {neighbor['fitness']:.4f}")
        avg_neighbor_fitness = sum(n['fitness'] for n in neighbor_analysis) / len(neighbor_analysis) if neighbor_analysis else 0
        best_fitness = self.evaluate_expression(self.best_expression)[0]
        print("\nInsights:")
        print(f"Average Neighbor Fitness: {avg_neighbor_fitness:.4f}")
        print(f"Best Expression Fitness: {best_fitness:.4f}")
        if best_fitness > avg_neighbor_fitness:
            print("The best expression outperforms its neighbors, suggesting it's a local optimum.")
        else:
            print("The best expression has similar performance to its neighbors, suggesting potential for further optimization.")
        self.plot_expression()

    def plot_expression(self):
        if not self.plot_results:
            return None
        try:
            expr = sp.sympify(self.best_expression)
            func = sp.lambdify(list(self.variables.values()), expr, modules=["numpy", {
                'cosh': self.safe_cosh, 'sinh': self.safe_sinh, 'sqrt': self.safe_sqrt,
                'Max': self.safe_max, 'Min': self.safe_min, 'Abs': np.abs,
                'ceiling': np.ceil, 'floor': np.floor
            }])
            data_values = [self.data[var].values for var in self.variables]
            predictions = func(*data_values)
            predictions = np.asarray(predictions)
```

```python
            if predictions.ndim == 0:
                predictions = np.full_like(self.target, predictions)
            elif predictions.ndim > 1:
                predictions = np.squeeze(predictions)
                print(f"Squeezed predictions shape: {predictions.shape}")
            if np.isnan(predictions).any() or np.isinf(predictions).any():
                print(f"Invalid predictions generated for expression: {self.best_expression}")
                return None
            if np.all(predictions == 1):
                mse_str = "N/A"
                rmse_str = "N/A"
                r2_str = "N/A"
            else:
                mse = mean_squared_error(self.target, predictions)
                rmse = np.sqrt(mse)
                r2 = r2_score(self.target, predictions)
                mse_str = f"{mse:.2f}"
                rmse_str = f"{rmse:.2f}"
                r2_str = f"{r2:.2f}"
            fig, ax = plt.subplots(figsize=(10, 6))
            ax.scatter(self.target, predictions, alpha=0.5, label='Predicted vs Actual')
            ax.plot([self.target.min(), self.target.max()], [self.target.min(), self.target.max()], 'r--', label='Ideal Fit')
            actual_func = self.actual_function
            predicted_func = sp.pretty(expr)
            ax.legend(title=f"Actual: {actual_func}\n"
                            f"Predicted: {predicted_func}\n"
                            f"MSE: {mse_str}, RMSE: {rmse_str}, R²: {r2_str}")
            ax.set_xlabel("Actual Values")
            ax.set_ylabel("Predicted Values")
            ax.set_title("Actual vs Predicted Values for Expression")
            plt.show()
            return fig
        except Exception as e:
            print(f"Error plotting expression: {self.best_expression}. Error: {str(e)}")
            return None

    def clear_caches(self):
        self._expression_depth_cache.clear()
        self._evaluate_expression_cache.clear()
        self._subexpressions_cache.clear()
        self._tree_edit_distance_cache.clear()
        self.similarity_cache.clear()
        self.lambdified_cache.clear()
```